\documentclass[sigconf,screen]{acmart}
\usepackage{pgfplots}
\usepackage{algorithm}
\usepackage{algorithmic}
\usepackage{listings}
\usetikzlibrary{arrows.meta, chains, positioning, shapes.symbols}
\usetikzlibrary{decorations,calligraphy}
\usepackage{numprint}
\usepackage{mathtools}
\usepackage{amsmath,mleftright}
\usepackage{multirow}
\usepackage{pifont}
\usepackage{colortbl}

\newtheorem{definition}{Definition}
\newcommand{\eat}[1]{}
\usepackage{balance}

\usepackage{tcolorbox}
\usepackage{caption}
\usepackage{subcaption}
\usepackage{multirow, graphicx}
\usepackage{rotating}
\usepackage{xcolor}
\usepackage{makecell}

\definecolor{codegreen}{rgb}{0,0.6,0}
\definecolor{codegray}{rgb}{0.5,0.5,0.5}
\definecolor{codepurple}{rgb}{0.58,0,0.82}
\definecolor{backcolour}{rgb}{0.95,0.95,0.92}

\lstdefinestyle{mystyle}{
    backgroundcolor=\color{backcolour},   
    commentstyle=\color{codegreen},
    keywordstyle=\normalsize\ttfamily,
    numberstyle=\tiny\color{codegray},
    stringstyle=\color{codepurple},
    basicstyle=\ttfamily\normalsize,
    breakatwhitespace=false,         
    breaklines=true,                 
    captionpos=b,                    
    keepspaces=true,                 
    numbers=left,                    
    numbersep=5pt,                  
    showspaces=false,                
    showstringspaces=false,
    showtabs=false,                  
    tabsize=4,
    stepnumber=1
}
\def\shorten{\looseness=-1}

\newcommand{\dbpedia}{DBpedia}
\newcommand{\sysName}{\textsc{Chatty-Gen}}
\newcommand{\myNum}[1]{(\emph{#1})}

\npstyleenglish

\AtBeginDocument{%
  }

\setcopyright{acmlicensed}
\copyrightyear{2025}
\acmYear{2025}
\acmDOI{10.1145/3709681}
\acmConference[]{}{}
\begin{document}

\title{Dialogue Benchmark Generation from Knowledge Graphs with Cost-Effective Retrieval-Augmented LLMs}

% \author{Paper-ID: 560} 892
% \author{Paper-ID: 892} 

% \renewcommand{\shortauthors}{Paper-ID: 892} 

\author{Reham Omar}
\affiliation{%
  \institution{Concordia University}
  \country{Canada}
}
\email{reham.omar@mail.concordia.ca}

\author{Omij Mangukiya}
\affiliation{%
  \institution{Concordia University}
  \country{Canada}
}
\email{omij.mangukiya@mail.concordia.ca}

\author{Essam Mansour}
\affiliation{%
  \institution{Concordia University} 
  \country{Canada}
}
\email{essam.mansour@concordia.ca}

\begin{abstract}
\sloppy
Dialogue benchmarks are crucial in training and evaluating chatbots engaging in domain-specific conversations. Knowledge graphs (KGs) represent semantically rich and well-organized data spanning various domains, such as DBLP, DBpedia, and YAGO. Traditionally, dialogue benchmarks have been manually created from documents, neglecting the potential of KGs in automating this process. Some question-answering benchmarks are automatically generated using extensive preprocessing from KGs, but they do not support dialogue generation. This paper introduces {\sysName}, a novel multi-stage retrieval-augmented generation platform for automatically generating high-quality dialogue benchmarks tailored to a specific domain using a KG. {\sysName} decomposes the generation process into manageable stages and uses assertion rules for automatic validation between stages. Our approach enables control over intermediate results to prevent time-consuming restarts due to hallucinations. It also reduces reliance on costly and more powerful commercial LLMs. {\sysName} eliminates upfront processing of the entire KG using efficient query-based retrieval to find representative subgraphs based on the dialogue context.
Our experiments with several real and large KGs demonstrate that {\sysName} significantly outperforms state-of-the-art systems and ensures consistent model and system performance across multiple LLMs of diverse capabilities, such as GPT-4o, Gemini 1.5, Llama 3, and Mistral.\shorten
\end{abstract}

\maketitle
\sloppy

\section{Introduction}

Conversational AI systems, like chatbots (e.g., ChatGPT~\cite{chatgpt} and Gemini~\cite{gemini}) and virtual assistants (e.g., Alexa and Siri), have become ubiquitous in offering general information and facilitate task completion~\cite{gpt4, gemini}. However, evaluating their performance within a specific domain remains a challenging task. Dialogue benchmarks provide a systematic way to assess these systems and improve the capabilities of general virtual assistants~\cite{p_assistant} in applications, such as 
educational chatbots~\cite{edu_chatbot}, 
onboarding new users to a specific domain~\cite{AttnIO}, and online tutoring~\cite{tutoring}.\shorten

\begin{figure}[t]
% \vspace*{-1ex}
  \centering
  \includegraphics[width=\columnwidth]{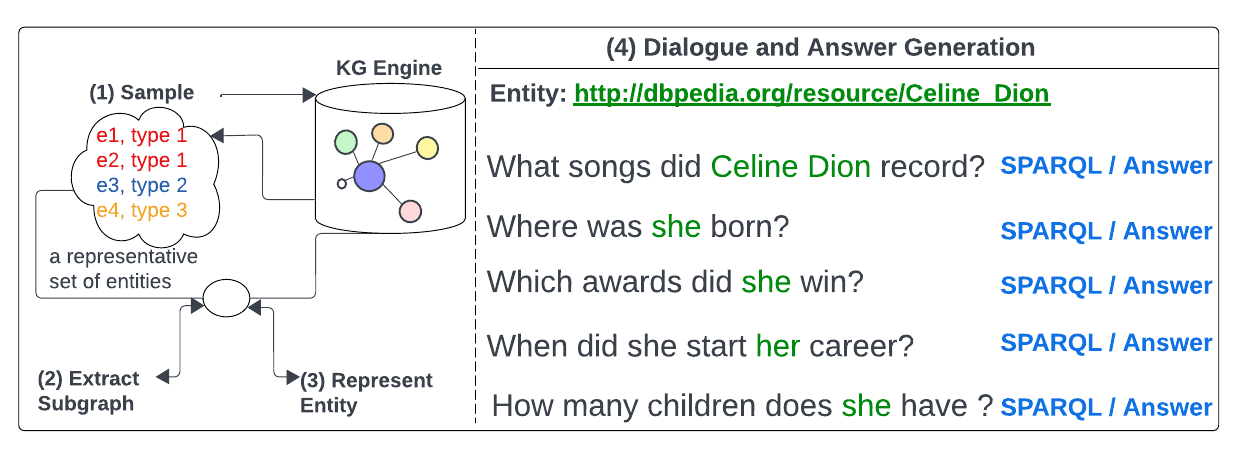}
  \vspace*{-4ex}
\caption{An illustration of the steps required to generate a dialogue from the entity "Celine Dion" in the DBpedia KG. The subgraph of Celine Dion serves as the dialogue context.}
\label{fig:example1}
% \vspace*{-3ex}
\end{figure}

Unlike unstructured datasets such as documents, knowledge graphs (KGs) represent semantically rich and organized data across various application domains. 
Examples include KGs about scientific publications, such as the Microsoft Academic Graph\footnote{\url{https://makg.org/}} and DBLP\footnote{\url{https://dblp.org/}}, and general-fact KGs like {\dbpedia}\footnote{\url{https://dbpedia.org/sparql}}, YAGO\footnote{\url{https://yago-knowledge.org/sparql}}, and Wikidata\footnote{\url{https://query.wikidata.org/}}.
A KG is a heterogeneous graph that includes nodes representing entities of various types, such as \textit{Person}, \textit{Creative Work}, or \textit{Place}, and edges representing relationships, such as \textit{birthPlace}, \textit{award}, or \textit{child}, between these entities. Both entities (vertices) and relationships (predicates) are represented using Uniform Resource Identifiers (URIs). Each entity can serve as a seed for a dialogue, with questions related to at least one predicate, as shown in \autoref{fig:example1}. For instance, \textit{Where was she born?} is a question related to the entity \textit{Celine Dion} and the predicate \textit{birthPlace}.
Hence, semantically rich KGs present a valuable resource for automating the generation of dialogue benchmarks for both general and domain-specific knowledge. \shorten

Automating dialogue benchmark generation from a KG presents unique challenges. \autoref{fig:example1} outlines the steps involved in generating a dialogue benchmark from the DBpedia KG as follows: 
\myNum{1} \textbf{Sample Entities}, which involves sampling representative entities from the KG. It is impractical to include all KG entities in the dialogue benchmark generation, as some KGs contain millions or billions of entities. Selecting a representative sample of key entities is crucial to achieving high-quality benchmarks in a reasonable time. Our example shows the entity \textit{Celine Dion} of type \textit{Person}.
\myNum{2} \textbf{Extract Context}, in which a subgraph surrounding the chosen entity is extracted to represent the dialogue context and ensure the dialogue remains focused and relevant. 
\myNum{3} \textbf{Extract Textual Entity Representation}, to generate a question about an entity, we need a human-readable representation. Using URIs (e.g. \url{http://dbpedia.org/resource/Celine_Dion}) within natural language questions is impractical, necessitating a more user-friendly representation (e.g. Celine Dion). Some URIS are encoded with no human-readable text. Hence, this is a challenging task.
\myNum{4} \textbf{Maintain Dialogue Flow}, which maintains the coherence of the conversation. The generated questions should build upon each other within the dialogue. For example, the primary entity is explicitly introduced once, with subsequent questions referring to the entity indirectly, as shown in \autoref{fig:example1}. 
\myNum{5} \textbf{Answer Generation:} Generating complete benchmarks requires answers for each question. This process involves utilizing structured queries, such as SPARQL. This necessitates formulating SPARQL queries corresponding to each question in the dialogue. 
\myNum{6} \textbf{KG Agnostic Design}, where the dialogue generation platform should be able to operate on any KG without requiring KG-specific modifications. This makes the platform useful across different domains and types of KGs.

\begin{table*}[t]
\vspace*{-3ex}
  \caption{A comparative analysis between question-answering and dialogue benchmark systems.}
  \vspace*{-2ex}
  \label{tab:back}
  \begin{tabular}{cccccc}
    \toprule
     \textbf{Benchmark} & \textbf{Node Selection} & \textbf{Entity Representation} & \textbf{Question Generation} & \textbf{Data Source} & \textbf{Dialogue}\\
    \midrule
    \textbf{CoQA\cite{coqa}} & N/A & N/A & Manual & Documents & \checkmark\\
    \textbf{QuAC\cite{quac}} & N/A & N/A & Manual & Documents & \checkmark\\
    \textbf{CSQA\cite{csqa}} & N/A & N/A & Template-based & KG & \checkmark \\
    \textbf{Head-to-Tail\cite{head_to_tail}} & Popularity + Random & N/A & Template-based & KG &x \\
    \textbf{Maestro\cite{maestro}} & Pre-processing + Popularity & User-defined & Rule-based & KG &x \\
    
    \midrule
    \textbf{{\sysName}} & Node Types Importance & Automatic & LLM-based & KG &\checkmark \\
  \bottomrule
\end{tabular}
\vspace*{-2ex}
\end{table*}

Large language models (LLMs), such as GPT~\cite{gpt4}, Gemini~\cite{gemini}, and Llama~\cite{llama}, hold significant potential for automating KG-based dialogue benchmark generation. 
However, traditional methods for generating dialogue benchmarks rely on manually created benchmarks from documents, such as Dream~\cite{dream}, MultiWOZ~\cite{multioz}, CoQA~\cite{coqa}, and QuAC~\cite{quac}. These manual approaches are labor-intensive, expensive, and not scalable.
Additionally, template-based systems exist for KG-based dialogue benchmarks, like CSQA~\cite{csqa}, and for question-answer benchmarks, i.e., standalone and self-contained questions, such as LCQuAD~\cite{lcquad,lcquad2} and Head-to-Tail~\cite{head_to_tail}. These template-based systems face similar issues and require template redesign for each new KG. 
Maestro~\cite{maestro} is a rule-based system automating the generation of question-answer benchmarks but does not support dialogue generation. 
Maestro faces challenges, including the need for extensive pre-processing, user input, and code adjustments for each new KG. Overall, these existing systems struggle to adapt to arbitrary KGs.
None of the existing systems leverages LLMs for KG-based dialogue benchmark generation, which presents additional challenges. One critical challenge is mitigating hallucinations\cite{hallucinate}—ensuring the generated dialogues strictly adhere to factual information within the KG. Another challenge lies in developing a versatile platform that can utilize various LLMs to perform the dialogue benchmark generation.\shorten

This paper introduces {\sysName}, a novel multi-stage Retrieval-Augmented Generation (RAG) platform for automatically generating high-quality dialogue benchmarks tailored to a specific domain using a KG. {\sysName} employs a multi-stage pipeline with zero-shot learning to simplify dialogue benchmark generation into manageable sub-tasks, each addressed by specifically crafted prompts without curated examples. 
{\sysName} utilizes assertion rules defined based on the expected output of each stage and leverages KG information to validate intermediate results automatically. Hence, {\sysName} mitigates hallucinations before proceeding to the next stage. 
% It also minimizes restarts to only problematic sub-tasks. 
Our multi-stage approach enables {\sysName} to accommodate and work efficiently with LLMs of diverse capabilities. {\sysName} utilizes an efficient query-based retrieval augmented technique to identify subgraphs with a large variety of predicates. Then, it summarizes these graphs to provide the prompt with rich yet condensed information as dialogue context. {\sysName} also predicts textual representations of the selected entities to generate human-like questions suitable for dialogue.
Additionally, {\sysName} automatically generates SPARQL queries for each question without prior knowledge of the KG. 

We evaluate {\sysName}'s performance on four diverse real-world KGs: DBpedia, Yago, and DBLP. Additionally, we assess {\sysName}'s compatibility with various LLMs, including commercial models like GPT-4o\cite{gpt4o} and Gemini-pro-1.5\cite{geminiteam2024gemini15}, as well as open-source models like Llama-3\cite{llama3modelcard} and CodeLlama\cite{codellama}. Our multi-stage approach allows {\sysName} to integrate with a single or multiple LLMs. We demonstrate that {\sysName} using multiple LLMs, such as Llama-3 and CodeLlama, achieves the same success rate, quality, and comparable processing time as {\sysName} with GPT-4o. Furthermore, {\sysName} exhibits significant time efficiency in end-to-end benchmark generation compared to Maestro, the current state-of-the-art system for standalone and self-contained questions. For large KGs like DBpedia, {\sysName} shows a remarkable time improvement of 99\%, reducing the process from 30 hours by Maestro to just 10 minutes by ours. Overall, {\sysName} offers a cost-effective and versatile solution for generating high-quality KG-based dialogue benchmarks within a reasonable timeframe.\shorten

In summary, our contributions are:
\begin{itemize}
    \item {\sysName}, the first fully automated RAG-based platform for generating a dialogue benchmark from a KG (Section 3).
    
    \item An effective and diverse context retrieval-augmented method for selecting representative seed entities with rich subgraphs from an arbitrary KG with minimal overhead (Section 4).
    
    \item A multi-stage RAG-based approach with automatic validation that mitigates hallucinations and achieves accurate dialogue benchmark generation while reducing the time and token consumption across different LLMs (Section 5).\shorten
    
    \item A comprehensive evaluation using four real KGs from different domains and various LLMs shows that {\sysName} significantly outperforms state-of-the-art systems in both quality and time performance. {\sysName} also works seamlessly with arbitrary KGs and ensures consistent performance across commercial and open-source LLMs (Section 6). 
    % including GPT-4o, Gemini 1.5, Llama 3, and Mistral (Section 6).
    
\end{itemize}

\vspace*{-1.5ex}
\section{Background: Limitations and Challenges}
This section reviews the capabilities of existing systems in generating dialogue benchmarks from KGs and highlights the challenges retrieval-augmented LLMs face in handling complex and knowledge-intensive tasks like dialogue benchmark generation.

\vspace*{-1.5ex}
\subsection{KG-based Dialogue Benchmarks}
A KG-based dialogue benchmark consists of dialogues, each containing questions along with their corresponding SPARQL queries to retrieve the answers from the KG. We define it as follows:
\vspace*{-1ex}
\begin{definition}
$D = \{e, KG, Q, SQ\}$ is a dialogue where:
\begin{itemize}
    \item \textbf{Seed Entity} $e$: The entity that the dialogue revolves around.
    \item \textbf{Questions} $Q$: An ordered list of questions asked during the dialogue, denoted as $Q = [Q_1, Q_2, \ldots, Q_n]$. The first question $Q_1$ must be a standalone and self-contained question, while subsequent questions ($Q_i$ for $i > 1$) depend on previous questions.\shorten 
    \item \textbf{SPARQL Queries} ($SQ$): A list of SPARQL queries, one for each question in $Q$, used to retrieve answers from $KG$.
\end{itemize}
\vspace*{-1ex}
\end{definition}

There are no existing systems that automatically generate dialogue benchmarks (DBen) from KGs. \autoref{tab:back} summarizes the capabilities of most relevant systems with focus on their support for dialogue and characteristics that might limit their compatibility with arbitrary KGs. 

CoQA~\cite{coqa} and QuAC~\cite{quac} are well-known manually generated DBen from documents. The context extraction involves selecting passages from source documents, where entities are embedded in the extracted passages in a human-readable format, i.e., no entity sampling or textual representation steps.\shorten

CSQA~\cite{csqa} is a semi-automatically generated DBen for the Wikidata KG. It utilizes a predicate-centric approach, which extracts all $330$ Wikidata predicates with their associated subjects and objects. In CSQA, standalone and self-contained questions are manually created. Annotators formulate these questions using the provided subject-predicate-object triples. To convert these questions into dialogues, CSQA employs pre-designed templates, which consist of connected question-answer pairs. Each pair shares a predicate or an entity. Finally, annotators manually adjust the generated dialogues to introduce complexities, such as coreferences, ellipses, incompleteness, and contextual dependence. Consequently, supporting new KGs with CSQA is a labor-intensive process. 

Head-to-Tail~\cite{head_to_tail} uses templates to create question-answering (QA) benchmarks from KGs. 
It uses a popularity-based approach for entity selection, where entities are sorted by popularity. Question generation focuses on attributes of randomly sampled entities and relies on manually created templates for each attribute. This necessitates defining numerous templates for various attribute-node type combinations, making it impractical for KGs with many entities. Generated questions often repeat for entities of the same type, resulting in low-quality dialogues if grouped. Supporting a new KG with Head-to-Tail requires designing a large number of templates to cover all attributes across node types.\shorten

\begin{figure*}[t]
\vspace*{-3.5ex}
  \centering
  \includegraphics[width=\textwidth]{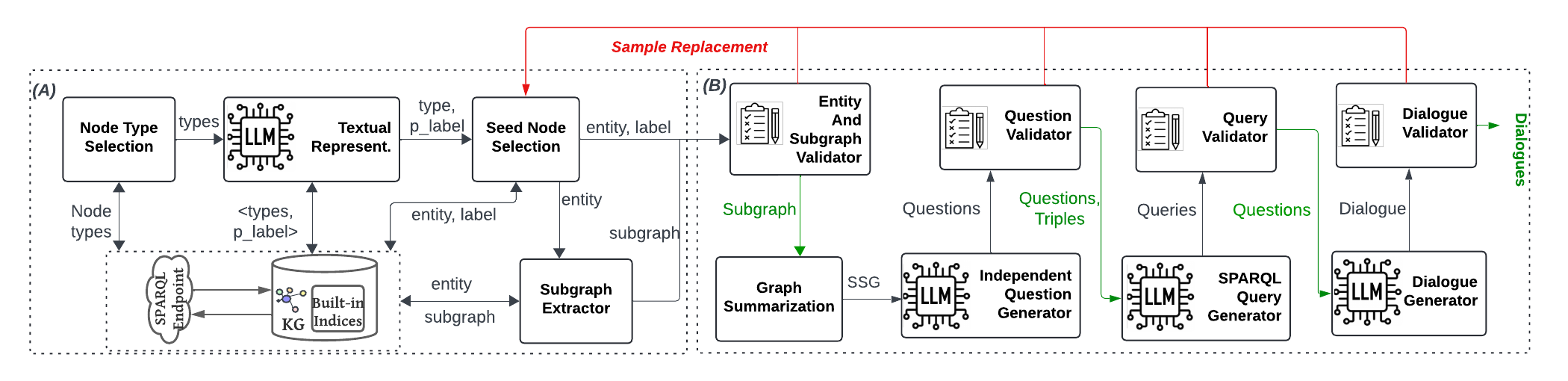}
  \vspace*{-8ex}
\caption{
{\sysName}'s architecture includes two main phases: A) \underline{Dialogue Context Extraction:} involves a node-type retrieval method to predict entity's textual representations from the KG and extracting seed entities with surrounding subgraphs as dialogue context. B) \underline{Dialogue Generation:}  employs three LLM-based steps: generating self-contained questions, formulating SPARQL queries from questions and triples, and organizing them into a coherent dialogue.
}
\label{fig:framework}
\vspace*{-3.5ex}
\end{figure*}

Maestro~\cite{maestro} is a rule-based system that automatically generates QA benchmarks from KGs with no support for dialogues. 
It begins with extensive preprocessing to construct a predicate lexicon table by extracting predicates, labels, and connected entity types from the KG, with processing time scaling to the number of predicates and edges. It selects entities from the lexicon table but doesn't account for node type distribution. Hence, Maestro fails to ensure coverage of all critical node types in the KG.
Maestro requires users to manually specify a single KG predicate for extracting textual entity representation, which is inappropriate for a KG with diverse node types. For instance, "name" for Person nodes and "title" for Publication nodes serve distinct purposes. Question generation relies on rules and templates for each subgraph shape, with a bias towards boolean questions, diverging from the informative questions in human dialogues.
Questions generated by Maestro cannot form coherent dialogues as all questions for a seed entity yield the same answer (the seed entity itself). While Maestro supports multiple KGs, integrating a new one requires about 600 lines of code, posing challenges for non-technical users.

\subsection{Retrieval-Augmented LLMs and Prompting}

Retrieval-Augmented Generation (RAG) has emerged as a powerful technique to address the primary challenges of LLMs in performing knowledge-intensive tasks that demand factual grounding and efficient knowledge acquisition~\cite{RAG-Survey, RAD-Ben, CRAG}, such as generating dialogue benchmarks from a KG. Retrieval-augmented LLMs utilize RAG mechanisms to enhance their performance in such tasks.  RAG retrieves relevant information and feeds this enriched context to LLMs. For instance, generating dialogue benchmarks from a KG involves integrating varied and precise information from the KG. Consider a chatbot consulting a KG of scientific publications (e.g., DBLP) to create dialogue benchmarks for academic research or a creative tool using a KG of artists and singers (e.g., DBpedia or YAGO) to develop detailed and informative dialogues about their lives and works. RAG offers a more flexible and data-efficient alternative over traditional fine-tuning methods~\cite{gpt3}. Fine-tuning requires significant computational resources, especially for massive models like GPT-4~\cite{gpt4} and LLAMA~\cite{llama}. Hence, it is a more common technique for relatively smaller models, such as BERT~\cite{bert} and T5~\cite{T5}.\shorten

RAG helps mitigate LLM hallucinations~\cite{hallucinate}, which occur when LLMs generate factually incorrect answers or outputs deviating from the given prompt~\cite{hallucinate}. Hallucinations stem from the unstructured nature of training data~\cite{graphqa} and the rare occurrences of certain information in the training data~\cite{hallucinate}. Examples include: \myNum{I} ambiguous dialogues, where the LLM starts with a context-lacking question, e.g., "What is his nationality?"; \myNum{II} non-existing facts, where the LLM generates questions that do not map to any facts (triples) in the KG; \myNum{III} incorrect SPARQL, where the LLM generates SPARQL queries that do not correspond to the questions. Overcoming hallucinations is crucial for ensuring the reliability and usability of dialogue benchmarks generated from a KG. 

RAG leverages in-context learning techniques like prompting~\cite{gpt3}, where clear instructions guide the LLM toward a specific task. Additionally, RAG can adapt to new scenarios through few-shot learning (using a few examples) or zero-shot learning (relying solely on prompts). The later establishes a highly customizable and versatile pipeline across unseen classes or situations, e.g., a new KG.
Several techniques aim to enhance prompt engineering~\cite{prompt_change}: \myNum{I} \textbf{Chain-of-Thought (CoT)}~\cite{cot} improves LLM reasoning by embedding intermediate steps within the prompt, guiding the model through problem-solving. While effective for tasks with clear steps, CoT does not seamlessly translate to all LLM architectures and often needs substantial adjustments to work effectively across different models. \myNum{II} \textbf{Zero-shot CoT}~\cite{zeroshot_cot} offers a variant of CoT that employs a dual-prompt strategy: the first prompts the LLM to extract reasoning, while the second provides both the question and the extracted reasoning for solving the task. \myNum{III} \textbf{Least-to-most prompting}~\cite{least_to_most} addresses tasks requiring generalization beyond the examples provided. It involves breaking down the task into simpler subtasks, which the LLM then solves sequentially until achieving the desired output. While effective for tasks like math reasoning~\cite{least_to_most}, it can be more costly and prone to error propagation in tasks like text-to-SQL~\cite{text_sql_cot}, which is similar to part of our task. \myNum{IV} \textbf{Automatic Prompting Engineering (APE)}~\cite{APE} automates prompt generation and selection to reduce human effort and testing. Using input-output samples, the LLM generates instructions evaluated on subset of data, iterating until convergence based on scores.\shorten

Generating dialogue benchmarks from KGs involves implicit reasoning about KG relationships, which traditional prompts struggle to capture as they focus on specific reasoning styles. Existing techniques are also not optimized for complex tasks that generate multiple outputs, such as dialogues paired with SPARQL queries in JSON format. Customizing prompts for such tasks can overwhelm some LLMs, limiting their efficiency. Therefore, these techniques often lack broad applicability across diverse LLM architectures. Achieving consistent performance across LLMs of varying capabilities remains a significant challenge in developing dialogue benchmark generation systems.
RAG also faces additional challenges, such as selecting representative seed entities from the KG and extracting rich yet concise subgraphs. Minimizing the latency of subgraph extraction is crucial. Furthermore, RAG must effectively synthesize information to generate dialogues with diverse questions while avoiding hallucinations. Another significant challenge is developing an automatic validation mechanism that uses the retrieved information to identify and mitigate hallucinations. 
These challenges open new research directions for developing a cost-effective and versatile dialogue benchmark system using retrieval-augmented LLMs.\shorten
%Addressing these challenges raises promising research directions for developing a cost-effective and versatile dialogue benchmark generation system based on retrieval-augmented LLMs. 

\vspace*{-2ex}
\section{The {\sysName} Platform Overview}
% {\color{blue}
{\sysName} automates the generation of domain-specific dialogue benchmarks for chatbot evaluation, which traditionally relies on human input for questions and templates. By leveraging KGs, it incorporates diverse key concepts and topics (node types and facts) to ensure comprehensive testing across various entity types. To overcome the limitations of traditional prompting methods, {\sysName} employs a cost-effective multi-stage pipeline that reduces both the financial cost of using LLMs and the processing time for large KGs.
% }\highlightedReply{}{R1.O1}\
%
{\sysName} includes two main phases: A) dialogue context extraction in the form of diverse subgraphs ($SG$) and B) dialogue generation for each $g \in SG$. The second phase is further broken down into four stages, where each stage's output is automatically validated before being used as input for the next stage. This ensures the correctness and consistency of the generated outputs. Validation is based on assertion rules defined by benchmark criteria and leverages KG information to mitigate potential LLM hallucinations.\shorten

The dialogue context extraction phase utilizes our efficient and diverse subgraph retrieval-augmented method. This method analyzes the KG node type distribution and selects a representative sample of seed nodes for each critical node type. KGs can contain millions or billions of nodes representing entities of ten to thousands of types. Moreover, our query-based retrieval efficiently analyzes these types by leveraging built-in RDF engine indices. Hence, our method reduces the preprocessing time and enhances overall efficiency. {\sysName} leverages LLMs to identify a textual label associated with a certain node type that can be used as an entity label. The textual label enables the LLM to generate better dialogues. {\sysName} selects nodes and their textual representations from the KG at runtime via SPARQL queries. These extracted entities are used to retrieve surrounding subgraphs, representing the context for question generation within the dialogue. LLMs have token limits, so {\sysName} utilizes constraints to select seed nodes with textual representations and subgraphs of specific lengths and sizes. This method ensures the extraction of subgraphs with diverse predicates, generating meaningful dialogues.

{\sysName} utilizes the extracted subgraph and textual representations through our four-stage pipeline to generate the dialogue. First, the subgraph is fed to a summarization module, which retains only information relevant to the task and eliminates potentially confusing details. Second, questions are generated based on the summarized subgraph. Each question is linked to the specific triple in the subgraph from which it was generated. Next, the questions and their corresponding triples are provided to the answer generation module. This module utilizes zero-shot learning and prompts to generate a SPARQL query for each question, retrieving the answer from the KG.
Finally, the dialogue generation module transforms the independent questions into a coherent dialogue, ensuring the first question remains independent as the starting point. The final benchmark dataset consists of three components: dialogue, independent questions, and corresponding SPARQL queries. {\sysName} avoids the need for complex, customized prompts used in techniques like CoT. Instead, it focuses on simpler prompts, enhancing its broad applicability across diverse LLM architectures. {\sysName} also regenerates only for specific subtasks, reducing overall costs.\shorten

\section{Dialogue Context Extraction as Subgraphs}

% {\color{blue}
The dialogue context extraction automatically identifies key concepts and topics (node types) in a KG, which may contain hundreds of types that do not represent the main concepts of a specific domain. This step is essential for creating comprehensive domain-specific dialogue benchmarks. Different entities of these node types vary in the density of surrounding facts (context), resulting in subgraphs of different sizes. Given that large KGs contain millions of entities, {\sysName} aims to identify a representative set of key entities with sufficient surrounding facts to generate diverse dialogues. For example, subgraphs with repeated predicates do not facilitate varied dialogue generation. Thus, {\sysName} optimizes both cost and processing time.
% }\highlightedReply{}{R1.O1}
The subgraph of a given seed entity is defined as follows:\shorten

\vspace*{-1.5ex}
\begin{definition}
\label{def:subgraph}
Given a $KG$ with a set of vertices $V$ and a set of predicates $P$, let $e$ be the target seed entity. The subgraph $SG(e) = \{T\}$ is a set of triples $T = \langle s, p, o \rangle$ such that $s = e$ or $o = e$, $p \in P$, $s \in V$, and $o \in V$ or a literal.
\end{definition}
\vspace*{-1.5ex}

% {\color{blue}
A brute-force solution typically begins by extracting all entities from the KG during a pre-processing step. Various approaches can then be used to process these entities, such as identifying head, torso, and tail entities, grouping them by their degree (number of incoming and outgoing predicates), or categorizing them based on the types of connected nodes. After categorization, entities are selected, and their corresponding subgraphs are generated. The complexity of these approaches is proportional to the number of entities in the KG, which can reach millions or even billions. As a result, these methods have significant drawbacks, including long processing times, high computational costs, and large storage requirements. This makes brute-force methods impractical for dialogue context extraction from large KGs.
% }\highlightedReply{}{R1.O1}

To efficiently extract context for dialogue generation, our approach leverages the node types in a KG through a three-step process: \myNum{i} relevant node type selection, \myNum{ii} textual entity representation, and \myNum{iii} seed node and subgraph extraction. In the first step, we focus on identifying pertinent KG node types, representing different concepts such as people or locations. Subsequent steps utilize this node-type information to extract textual representations of entities and sample a representative set of entities for subgraph extraction and dialogue generation. This streamlined approach operates effectively with large KGs, eliminating the need for costly preprocessing that traverses the entire graph and stores all entities.\shorten

% \vspace*{7ex}
\subsection{Identifying Representative Node Types}
\label{sec:seed}

A KG contains nodes representing real-world entities of different types based on the application domain. For example, in the DBLP academic graph, \url{https://dblp.org/rdf/schema#Publication}is a domain-specific type representing scholarly publications. There are also metadata types that describe properties of the data within the KG and are not representative of core entities. Examples include \url{http://www.w3.org/2002/07/owl#DatatypeProperty} and RDF engine-specific types like \url{http://www.openlinksw.com/schemas/virtrdf#array-of-string}. Due to their nature, metadata types are less helpful in generating dialogues.
% {\color{blue} 
Within domain-specific types, there exist rare types and shadowed types. Rare types are those represented by a very low percentage of entities within the KG, indicating they are less critical to KG topics and do not represent the main focus of the KG.  Examples include \url{http://schema.org/Reservoir} in Yago KG, which represents only 0.020\% of the KG's entities. Shadowed types are parent types whose entities mainly belong to a single child type. For instance, in DBLP KG, \url{https://dblp.org/rdf/schema#Creator} is a parent of \url{https://dblp.org/rdf/schema#Person}, and over 99\% of Creator entities belong to Person type. Including both parent and child types would introduce redundancy in the final dialogues. 
To address this, {\sysName} allows users to set thresholds to exclude rare and shadowed types. This enhances efficiency and reduces benchmark generation time.
% }\highlightedReply{}{R1.O2, C5}

This module identifies the most relevant node types within the KG's domain. Initially, we retrieve all node types from the KG and filter out metadata and other irrelevant types. Our goal is to pinpoint types that support the generation of coherent dialogues about concepts pertinent to the KG's specific domain. This strategy enhances time efficiency compared to brute-force methods because the number of node types in a KG is typically much smaller than the total number of nodes. Subsequently, we establish a distribution of entities to sample from each identified type.

\begin{algorithm}[t]
\vspace*{2ex}
\caption{Node Type Selection}
\label{alg:node_selection}
\begin{flushleft}
\textbf{Input:} $endpoint$: SPARQL endpoint, $m$: number of Dialogues, $domain$: prefixes of KG, $R$: Rare types threshold, $S$: Shadowed parents threshold\\
% \textbf{Output:} $dist$: Map of node type to the number of entities to sample
\textbf{Output:} $dist$: A map of node type to the number of entities

\end{flushleft}
\begin{algorithmic}[1]
% Get types, filter, and calculate percentage
\STATE $dist, type\_ratio \gets \{\}$
\STATE $types, count \gets getKGNodeTypes(endpoint, domain)$
\STATE $total \gets Sum(count)$
\FOR{every $\langle t, c \rangle \in \langle types, count \rangle$}
        \STATE $type\_ratio[t] \gets c / total$
\ENDFOR
\STATE $type\_ratio \gets removeRareTypes(type\_ratio, R)$
\STATE $type\_ratio \gets removeShadowedTypes(type\_ratio, S)$
% \STATE $types \gets removeRareTypes(types, r_{threshold})$
% \STATE $types \gets removeShadowedTypes(types, s_{threshold})$
% \STATE $ dist \gets getTypesDistribution(types, m)$
\FOR{every $\langle t, ratio \rangle \in type\_ratio$}
        \STATE $dist[t] \gets ratio * m$
\ENDFOR
\STATE $\textbf{Return dist}$
\end{algorithmic}
\vspace*{2ex}
\end{algorithm}

Algorithm \autoref{alg:node_selection} outlines the procedure for selecting node types and determining the number of entities to sample from each type, where each entity corresponds to a single generated dialogue. The algorithm takes five inputs: \myNum{1} KG SPARQL endpoint for accessing the KG, \myNum{2} the required number of dialogues to generate \( (m) \), \myNum{3} domain prefixes for identifying relevant KG types, \myNum{4} a rare type threshold \( (R) \) to filter out types with low entity counts, and \myNum{5} a shadowed parent threshold \( (S) \) to identify redundant types.
% {\color{blue}
Line 2 retrieves relevant types to the KG domain using the provided SPARQL endpoint and domain prefixes. 
We employ a count SPARQL query leveraging the predicate \textit{rdf:type}, which is indexed in RDF engines. This approach allows {\sysName} to efficiently obtain a complete list of KG types along with their corresponding counts.
Lines 3 to 6 calculate the percentage of each type in the KG based on these counts.  To focus on significant types, line 7 removes rare types with percentages below $R$. If $R=0$, all types are included. Line 8 excludes shadowed types by identifying parent-child relationships using SPARQL queries leveraging subject and predicate indices. 
Then, it removes parents if any of its children constitute more than $S$ of their total instances. If  $S=1$, all types are included. Finally, we recalculate percentages to reflect the reduced set of types. Lines 9 to 11 determine the number of samples required from each type. To ensure the sample distribution reflects the overall entity distribution in the KG, we utilize the percentages of each type along with the desired total number of entities, denoted as $(m)$.
% }\highlightedReply{}{R1.O2, C3, C4, C10}

\textbf{Complexity:} Algorithm \autoref{alg:node_selection} has a time complexity of \( O(q_{cost} + \tau) \), where \( \tau \) represents the number of node types and \( q_{cost} \) denotes the cost of executing the SPARQL query to retrieve types and their entity counts. The primary operations contributing to this complexity include retrieving all node types (\( q_{cost} \)) and iterating through them (\( \tau \)) for filtering and calculations.
The cost of retrieving node types (\( q_{cost} \)) varies based on the RDF engine used, leveraging its implementation and efficiency in handling queries with built-in indices. In a large KG, the number of nodes can reach millions or billions, while the number of distinct node types (\( \tau \)) is typically much smaller, ranging from tens to thousands. Hence, this approach is significantly more efficient compared to a brute-force solution that would necessitate iterating through all nodes in the KG, which would be computationally prohibitive for large-scale KGs.

\vspace*{-1.5ex}
\subsection{Textual Entity representation}
\label{sec:entity_rep}
The textual representation of entities significantly influences human comprehension and LLMs' ability to generate questions. KGs typically represent entities using URIs, which are not suitable for natural language processing tasks. Therefore, converting these URIs into human-readable text is crucial for question generation. One straightforward approach is to extract the final segment of the URI as the entity label (e.g., \url{http://dbpedia.org/resource/The_Future_of_Freedom_Conference}). However, this method is limited because many URIs contain IDs or alphanumeric combinations in their final segments (e.g., \url{https://dblp.org/pid/00/10071} or \url{https://dblp.org/rec/conf/valuetools/Coppa14}), making it difficult for LLMs to derive meaningful questions from them. Due to these limitations, a more effective strategy involves leveraging the entity's context within the KG. This context includes the predicates connected to the entity, which can provide better textual representations.\shorten 

Identifying a single universally applicable predicate across different KGs or even within different node types in a single KG is challenging. For example, Person entities may be connected with a "name" predicate, while Publication entities might use a "title" predicate.
To address this challenge, users with domain expertise can manually define mappings between node types and suitable predicates for retrieving entity labels. While feasible for smaller KGs with fewer node types, this manual approach becomes impractical for large-scale KGs like DBpedia or Yago, which contain hundreds or thousands of node types. Therefore, there is a need for an automatic method to identify the most appropriate predicate, referred to in this paper as the \textit{entity label}, for each node type within a KG.\shorten

We propose a novel approach using LLMs to automatically extract the entity label (\( e_L \)) for each node type within a KG. For a given node type (\( t \)), our method begins by randomly sampling an entity (\( e \)) belonging to that type from the KG. Next, we extract all outgoing predicates (\( P_{list} \)) connected to \( e \). We define this subtask with the task instruction ($I_{EL}$) to be sent to an LLM, as follows:\shorten
\begin{equation}
    \label{eq:entity}
    p_l = f(t, P_{list}, I_{EL})
\end{equation} 

To ensure the extracted representation is suitable for use within questions, we filter \( P_{list} \) to retain only predicates connecting \( e \) to string literals. This filtering focuses exclusively on predicates relevant to the entity's type, irrespective of the literal's length.
Instruction $I_{EL}$ utilizes zero-shot prompting to instruct the LLM to select the most representative predicate from the filtered list ($P_{list}$). The chosen predicate then becomes the entity label for the node type ($t$). This approach generates human-readable entity representations for all node types, regardless of the information encoded in the URIs. Additionally, it eliminates the need for technical knowledge to manually create the mapping between node types and entity labels. This improves the quality of questions asked and enhances the overall system usability.\shorten

%%%%DO NOT REMOVE IT, we may need it in the CRV
% {\color{orange}
The length of labels can vary significantly among entities within the same node type. For instance, consider the "Place" node type in the Yago KG: one entity might have the label "\textit{Administrative Department of Science, Technology, and Innovation}," while another might simply be "\textit{Admaston}." Hence, specific conditions regarding label usage are deferred to the entity-level processing stage.
% }

\begin{algorithm}[t]
\caption{Entity and Subgraph Extraction For Node Type $t$}
\label{alg:seed_subgraph_extraction}
\begin{flushleft}
\textbf{Input:} $endpoint$: SPARQL endpoint, $p_l$: entity label of $t$, $n$: number of entities of $t$, $BZ$: entity batch size, $h$: number of hops for subgraph,  $shape$: Subgraph Shape, $d$: Direction of predicates\\
% $label_{max}$: Maximum label length
\textbf{Output:} $Entity\_Subgraph$: A list of n valid entity, subgraph pair
\end{flushleft}

\begin{algorithmic}[1]
\STATE $entity\_subgraph \gets \{\}$
\WHILE{$entity\_subgraph.length < n$}
\STATE $e, e_L \gets getEntity(endpoint, p_l, BZ)$
% \If{$e_{label}.length > label_{max}$ }
% \State $\textbf{continue}$
% \EndIf
\STATE $g_{size} \gets getGraphSize(endpoint, e, h)$
\STATE $preds \gets countUniquePredicates(endpoint, e, h)$
\STATE $context_{valid} \gets validateContext(g_{size}, preds)$
% \If{$context_{valid}$ is $False$}
% \STATE $\textbf{continue}$
% \EndIf
\STATE $subgraph \gets extractSubgraph(e, endpoint, h, shape, d)$
\STATE $subgraph_{filtered} \gets filterSubgraph(subgraph, p_l)$
\STATE $subgraph_{valid} \gets validateSubgraph(subgraph_{filtered})$
\STATE $entity\_subgraph[e] \gets subgraph_{filtered}$
% \If{$subgraph_{valid}$ is $True$}
% \STATE $entity\_subgraph[e] \gets subgraph_{filtered}$
% \EndIf
\ENDWHILE
\STATE $\textbf{Return entity\_Subgraph}$
\end{algorithmic}
\end{algorithm}

\vspace*{-1.5ex}
\subsection{Seed Nodes and Subgraph Extraction}
This module efficiently extracts seed entities and their corresponding subgraphs from a KG, which serves as the context for dialogue generation. It leverages the node-type distribution and utilizes entity labels for each type. To optimize LLM processing, the extracted entities and subgraphs must meet specific criteria.
% {\color{blue}  
{\sysName} opts to select entities with concise labels to optimize performance and prevent LLMs from being overwhelmed by irrelevant details. For example, in the DBLP KG, a label like "\textit{Algorithmic Number Theory, First International Symposium, ANTS-I, Ithaca, NY, USA, May 6-9, 1994, Proceedings}" contains distracting elements such as dates and locations. While using LLMs to shorten labels may seem effective, it can alter the entity's meaning. Removing parts of a paper's title could confuse users or systems searching for a specific paper. We classify entities as valid if their labels fall below a predefined length threshold.
% }\highlightedReply{}{R1.O2, C6}
Secondly, the subgraph's size must be manageable within the LLM's context length. It also should feature an adequate number of unique predicates connecting entities to diverse facts. This diversity ensures the foundation for generating multiple distinct questions within a dialogue; typically, a minimum of three questions per dialogue is required. Subgraphs meeting these criteria are marked as valid. This approach offers efficiency gains over brute-force methods by avoiding storing all KG entities. Instead, it dynamically retrieves the necessary entities and their subgraphs directly from the KG endpoint without full graph traversal, thereby streamlining the process. Our technique prioritizes efficiency and guarantees a sufficient number of unique predicates to generate human-like and useful dialogues.\shorten

Algorithm \autoref{alg:seed_subgraph_extraction} outlines the process for extracting entity-subgraph pairs specific to a node type. This iterative process continues until the desired number of pairs ($n$) for the given node type $t$ are obtained. This procedure is repeated for each node type in the KG.
In line 3, the algorithm
% {\color{blue}
randomly
% } 
retrieves an entity of type $t$ along with its label using the entity label $e_L$. It ensures that the extracted entity's label length is suitable for processing by the LLM. 
% {\color{blue}
The algorithm retrieves entities in batches of size $BZ$ from the SPARQL endpoint. Leveraging the RDF predicate (\textit{rdf:type}) and object ($t$) indices, and batching entities helps optimize efficiency and minimize query overhead.
% }\highlightedReply{}{R1.O2}
Each entity is processed individually, and a new SPARQL query is issued only if the entire batch is exhausted before reaching the required number of pairs ($n$). For entities with suitable labels, lines 4-5 execute count queries to determine the graph size and the number of unique predicates within a specified hop limit ($h$) around the entity. 
% {\color{blue} 
In both queries, the RDF subject index is used for faster execution, since the subject is the entity. Moreover, they are count queries so they only return the counts, avoiding the overhead of retrieving and processing actual data, contributing to faster response times.
% }\highlightedReply{}{R1.O2}
Line 6 verifies that the graph size and the count of unique predicates meet the LLM processing constraints, providing sufficient information to generate diverse questions in a dialogue. Once the LLM's requirements are satisfied, subgraph extraction proceeds. Line 7 extracts the subgraph surrounding the seed entity based on predefined parameters, including the shape of the subgraph, the number of hops to traverse outward from the entity, and the direction of predicates (e.g., outgoing, incoming, or both directions). Line 8 filters out triples from the extracted subgraph based on several criteria.\shorten

Triples containing the entity label itself are excluded, as questions about these would yield the entity as the answer. Additionally, triples with predicates irrelevant to the KG's domain, such as \url{http://www.w3.org/1999/02/22-rdf-syntax-ns#type}, are removed as they lack specific information for generating meaningful dialogue questions. Furthermore, the algorithm filters out triples associated with excessively long string literals as objects. These literals, such as those connected via predicates like \url{http://dbpedia.org/ontology/abstract}, can unnecessarily increase the subgraph size and potentially confuse the LLM by leading to questions about the content of the objects rather than questions relevant to the entity itself. Similarly, predicates like \url{http://dbpedia.org/ontology/thumbnail}, which connect entities to image URLs, are excluded as they do not contribute to human-like dialogues. Line 9 validates the filtered subgraph to ensure it contains sufficient information (number of triples) required for dialogue generation. Valid subgraphs meeting the criteria are added to the final list of entity-subgraph pairs. If the subgraph does not meet the validation criteria, the algorithm proceeds to process the following entity. The algorithm's output is a list of valid entity-subgraph pairs extracted for each node type within the KG.\shorten

\textbf{Complexity:} 
Algorithm \autoref{alg:seed_subgraph_extraction} operates with a time complexity of \( O(\tau \cdot (q2_{cost} + K)) \), where: \myNum{i} \( \tau \) represents the number of selected node types as determined by Algorithm \autoref{alg:node_selection}.
    \myNum{ii} \( q2_{cost} \) denotes the cost associated with extracting the subgraph within the specified hop limit and direction from the RDF database.
    \myNum{iii} \( K \) signifies the number of triples within the extracted subgraph. The cost \( q2_{cost} \) varies depending on the RDF engine's implementation and the efficiency of its built-in indices. This approach achieves notable efficiency gains compared to a brute-force solution by minimizing the number of iterations (where \( \tau \ll V \), and \( V \) is the total number of entities in the KG) and avoiding the need for entity storage. Consequently, the algorithm operates with constant space complexity \( O(1) \).\shorten

\vspace*{-1.3ex}
\section{Our Multi-stage Dialogue Generation}
This section discusses generating dialogue benchmarks from subgraphs of seed entities and their labels.
We introduce the single-prompt approach, which involves processing each dialogue context through a single complex prompt. Then, we present our multi-stage approach using multiple distinct and manageable prompts
~\footnote{The details of all our prompts are provided in \href
{https://github.com/CoDS-GCS/Chatty-Gen/blob/main/Supplementary_material.pdf}
% {https://gitfront.io/r/ABC/cJkrczBNnLSc/Chatbot-Resources/blob/Supplementary_material.pdf}
{our supplementary materials}.}.

% {\textcolor{blue}{here}}.}.

\vspace*{-1.5ex}
\subsection{The Single-prompt Approach}
The single-prompt approach is conceptually straightforward and relatively easy to implement. It requires processing the subgraph only once, which seems computationally efficient. However, the single prompt typically encompasses several subtasks. First, the LLM must analyze the subgraph ($SG_{ser}$) with the entity label ($e_L$) to identify entities, relationships, and relevant information for dialogue generation. Based on this context, the LLM formulates a dialogue (\texttt{D}) of sequence of standalone questions (\texttt{Q'}) to explore the subgraph's relationships and properties. The LLM then generates corresponding SPARQL queries (\texttt{SQ}) to extract the answer from the subgraph. Then it weaves the generated questions and retrieved answers into a coherent dialogue reflecting the subgraph's underlying relationships. We define this task generating $n_q$ questions per dialogue using our instruction $I_s$ as follows:\shorten

\begin{equation}
\label{eq: q_gen}
    D, Q', SQ  = f(SG_{urls}, e_L, n_q, I_s)    
\end{equation}

In the prompt, we provide all the necessary information in one instruction. So, it makes the design crucial to guide the LLM effectively. However, this approach faces limitations despite its simplicity. Processing the entire dialogue context through a single prompt ($I_s$) can be overwhelming for less powerful LLMs, demanding high capabilities that may hinder broader applicability.
In a zero-shot setting, where no specific examples guide the LLM, it might struggle with the task's complexity.  That leads to hallucinations, such as irrelevant outputs, duplicate questions, or unclear formulations. Adding relevant examples to the prompt (few-shot learning) can mitigate these issues but may result in repetitive and predictable dialogues that lack diversity. Designing effective few-shot prompts requires a comprehensive set of diverse examples, which can be time-consuming and resource-intensive.
Thus, while the single-prompt approach serves as a baseline for dialogue benchmark generation and helps understand the challenges involved, its limitations highlight the need for more advanced methods.

\subsection{The Multi-stages Approach}
\label{sec:workflow}

Given the limitations of the single-prompt approach, we propose a multi-stage approach with assertion-based validations. This approach serializes the subgraph and leverages the LLM's text generation capabilities to generate dialogues across three stages:
\myNum{1} Standalone and self-contained (independent) question  generation: Generates questions about different facts in the KG.
\myNum{2} SPARQL query generation: Translates questions into SPARQL queries to extract answers from the KG.
\myNum{3} Dialogue generation: Transforms the independent questions into a coherent dialogue.
This breakdown enables better control over the process by validating intermediate results and guiding the LLM toward generating high-quality dialogues. While the multi-stage approach may incur slightly higher costs due to multiple prompts, it ensures that the overall dialogue generation process remains uninterrupted and avoids starting over due to hallucinations. This effectively addresses the limitations of the single-step approach and results in a more cost-effective method.\shorten

\textbf{Independent Question Generation:}
{\sysName} begins by generating a set of independent questions. For a given entity $e$, its corresponding serialized subgraph $SG_{ser}$, and the number of questions per dialogue $n_q$, we prompt the LLM to automatically generate a list $Q'$ of $n_q$ independent questions. Each question in $Q'$ must be answerable by the subgraph $SG_{ser}$. Moreover, answering questions in $Q’$ requires no context or knowledge from other questions.
% {\color{blue}
{\sysName} prioritizes a diversity of factual question types, such as "What," "How," and "When," over command-based questions like "Show," "Mention," and "Find." In the literature, factual questions have been more widely adopted in dialogue benchmarks compared to command questions. Examples of such conversational benchmarks include CoQA\cite{coqa}, QuAC\cite{quac}, CSQA\cite{csqa}, ConvQuestions\cite{convquestions}\footnote{The distribution of question types is shown in \href
{https://github.com/CoDS-GCS/Chatty-Gen/blob/main/Supplementary_material.pdf}
% {https://gitfront.io/r/ABC/cJkrczBNnLSc/Chatbot-Resources/blob/Supplementary_material.pdf}
{our supplementary materials}.}, and ConvMix\cite{convmix}.
% }\highlightedReply{}{R2.O2}
We define this subtask with our prompt instruction ($I_{IQ}$) as follows:\shorten
\begin{equation}
\label{eq: q_gen}
    Q' = f(SG_{ser},  e_L, n_q, I_{IQ})    
\end{equation}

\textbf{SPARQL Query Generation:}
The second stage is generating SPARQL queries for each independent question in $Q'$. Providing the LLM with only the question is insufficient for generating a correct SPARQL query. Without understanding the underlying structure (i.e., the subgraph), the LLM might introduce incorrect prefixes or use predicates that do not exist in the subgraph based on its interpretation of the question. To generate correct queries, the LLM needs the relevant context from the KG.
One approach is to inject the entire KG ontology into the LLM prompt, but this is computationally expensive and burdens the LLM with unnecessary information. A more efficient approach leverages the subgraph used to generate the questions by injecting only this subgraph into the prompt. However, even the subgraph might contain unrelated facts that could confuse the LLM. Therefore, we opted to provide the LLM with the minimal information necessary: the exact triples used to generate each question. To achieve this, we extended the independent question generation module to return these triples along with each question. This modification necessitates revising \autoref{eq: q_gen} to return the triples:
\begin{equation}
\label{eq:ssg_gen}
    Q', T_q = f(SG_{ser}, e_L, n_q, I'_{IQ})
\end{equation}

We leverage LLMs to generate SPARQL queries that answers the question from the KG given the question-triple pairs. The LLM relies solely on the question and its associated triples. This module output $SQ$ a list of SPARQL queries corresponding to the independent question list $Q’$. This approach generates syntactically correct SPARQL queries without relying on the KG ontology, as follows:
\begin{equation}
    \label{eq:sparql}
    SQ = f(Q', T_q, I_{SQ}) 
\end{equation}

\textbf{Dialogue Generation:}
The final stage of our workflow transforms the set of independent questions $Q'$ into a dialogue. This transformation must preserve the correspondence between each question and its corresponding SPARQL query. The first question remains independent, serving as a clear starting point of the conversation that requires no prior context. Subsequent questions build upon the answers or context provided by earlier questions. These specifications mirror the natural flow of human conversation, where each question is based on the established context.
We developed a four-step method that significantly improves the transformation of independent questions into dialogues, as follows:  \myNum{i} identifying the common entity shared across all questions, \myNum{ii} predicting the appropriate pronoun to replace the entity, \myNum{iii} substituting the entity with the chosen pronoun in each question, and \myNum{iv} ensuring the modified questions are grammatically and linguistically correct. 

Successful dialogue generation requires the LLM to correctly execute each sequential step. However, the complexity of this task can cause the LLM to mistakenly identify target entities for individual questions rather than recognizing a common target entity for the entire dialogue. This results in disconnected questions. For example, given the questions "\textit{What is the primary affiliation of Anne Condon?}" and "\textit{What is the nationality of Anne Condon?}", the LLM might incorrectly assign "\textit{primary affiliation of Anne Condon}" and "\textit{nationality of Anne Condon}" as target entities, respectively. To produce a coherent dialogue, the LLM needs to identify "\textit{Anne Condon}" as the common entity. To address this, we enrich the LLM prompt by including the entity label ($e_L$) as additional input as follows:\shorten

\begin{equation}
    \label{eq:dialogue}
    D = f(Q', e_L, I_{DG})
\end{equation}
This simplifies the task for the LLM. The LLM predicts the appropriate pronoun based on its understanding, then replaces the seed entity with this pronoun and makes necessary adjustments to ensure grammatically and linguistically correct questions. 

\subsection{Assertion-based validation}
\label{sec:validation}
To mitigate the potential for LLM hallucinations, we integrate a validator module into each stage of the workflow, as shown in Figure~\ref{fig:framework}. This module verifies that the LLM's outputs adhere to specified instructions. If an output is deemed invalid, we implement a retry mechanism with a maximum of three attempts. During each attempt, the LLM receives the same prompt and inputs. If a valid output is not obtained after three trials, the current seed entity is skipped, and we sample a new entity of the same node type. Our validation approach relies on assertions to ensure that the LLM's output satisfies specific constraints
~\footnote{The details of these constraints are available at \url
{https://github.com/CoDS-GCS/Chatty-Gen}
% {https://gitfront.io/r/ABC/cJkrczBNnLSc/Chatbot-Resources/}
}.
% {\color{blue} 
The assertion rules in our approach are independent of the KG and are prompt-related, based on the criteria of our multi-stage input/output process. This distinguishes our approach from template-based approaches that necessitate redesigning templates for each new KG.
% }\highlightedReply{}{C2} \shorten

\textbf{Question validator:} 
This module validates both questions and triples. We need to ensure that questions are independent, i.e., they explicitly mention the entity. This implies that the entity was not replaced by pronouns or general concepts that make the question context-dependent. For triples, we ensure that they exist within the provided subgraph. This guarantees the correctness of triples used for subsequent query generation, preventing the LLM from being confused by inaccurate information.

\textbf{Query validator:} 
Validation of generated queries ensures they are syntactically correct and capable of retrieving results consistent with the subgraph information. Only query lists containing at least three correct queries advance to the next step; otherwise, a new seed entity is selected.\shorten

\textbf{Dialogue validator:}
In validating dialogues, we assess the opening question for its ability to initiate the conversation independently. The second-to-last question is evaluated for its reliance on established context, ensuring it does not explicitly mention the target entity. Additionally, we verify that questions contain multiple words, as an LLM may generate a question of a single word like \textit{his?} instead of a full question. These responses are deemed invalid.

\vspace*{-1.5ex}
\subsection{Subgraph Serialization}
\label{sec:serialize}

Serialization involves encoding methods designed for nodes and predicates within the subgraph's triples. We utilize the textual representation extracted using the entity label (explained in \autoref{sec:entity_rep}) for node encoding. For predicate encoding, we remove unnecessary namespace information, allowing the LLM to focus on the meaning of the predicate. The final serialized subgraph string is a list of triples, where each node and predicate is encoded using its respective technique. We evaluated two serialization alternatives: the full subgraph serialization, which includes the full list of triples (subject, predicate, and object), and the summarized serialization, which includes only specific parts of the triples. This approach helps the LLM focus on question and query generation.

Serializing the full subgraph, which includes all its triples, presents several challenges: \myNum{1} Popular entities surrounded by many predicates can exceed the LLM's context length, leading to overflow. \myNum{2} Processing long subgraphs increases execution costs. \myNum{3} Including the answer in the triples (if the seed entity is the subject) can result in overly strict or incorrect SPARQL queries in some cases.
To address these issues, we employ summarized serialization, which utilizes modified triples instead of the complete set. This approach reduces the size of the input subgraph, allowing larger subgraphs to fit within the LLM's context length. It also decreases costs and improves token efficiency. By removing the object from the triples, the LLM can focus on aligning the seed entity and predicate to generate accurate SPARQL queries. Therefore, summarized serialization effectively mitigates the generation of incorrect queries.\shorten

Our summarization approach is driven by the observation that subgraphs often include numerous predicates, some of which appear multiple times, while others are unique. Repeated predicates provide valuable information for exploring the graph structure but contribute less to diverse question generation. For instance, consider a person entity who authored multiple publications. The subgraph may contain multiple triples such as $\langle$Author, authored, Paper$\rangle$. By leveraging just one of these triples, we can generate questions like \textit{What did the author publish?} or \textit{How many papers did she author?} by only using one triple. We define the summarized subgraph as follows:
\vspace*{-2ex}
\begin{definition}
\label{def:sum_subgraph}
Given a subgraph $SG(e) = \{T\}$, A summarized subgraph $SSG = \{T' ; T' \in SG \}$. For triple $T \in SG$, the corresponding $T' = \langle e, p, None \rangle$ or  $T' = \langle None, p, e \rangle$. Each $p$ exists once in $SSG$.\shorten
\end{definition}
\vspace*{-2ex}

\begin{algorithm}[t]
\caption{Summarizing Subgraph Algorithm}
\label{alg:summarize}
\begin{flushleft}
\textbf{Input:} $SG$: Subgraph \\
\textbf{Output:} $SSG$: Summarized Subgraph
\end{flushleft}

\begin{algorithmic}[1]
\STATE $p_u \gets getUniquePredicates(SG)$
\STATE $preds\_to\_triples \gets groupTriples(SG, p_u)$
\STATE $SSG \gets []$
\FOR{every $\langle p, t_{list}\rangle \in \langle preds\_to\_triples\rangle  $}
\STATE $t' \gets modifyTriple(e, t_{list}[0])$
\STATE $SSG \gets ssg \cup t'$
\ENDFOR
\STATE \textbf{Return} SSG
\end{algorithmic}
\end{algorithm}

Algorithm \autoref{alg:summarize} explains the summarization. Line $1$ extracts all unique predicates from the subgraph $SG$ to ensure complete fact inclusion. Each unique predicate represents a fact connected to the seed entity. Line $2$ groups the triples within the subgraph by predicate. This means all triples in a group share the same predicate. In lines $4-7$ for each (predicate, triple list) pair, the first triple is chosen as the group representative. Then we modify the triple based on the seed entity’s position within the triple. If the seed entity is the subject, the object is removed and vice versa. Then we add the modified triple to the summarized subgraph $SSG$. We provide the LLM with $SG_{ser}$, the string containing the list of modified triples from $SSG$.\shorten

\vspace*{-1ex}
\section{Evaluation}

% {\color{blue}
\subsection{{\sysName} Flexible Configurations}

{\sysName} is a flexible framework that allows users to generate benchmarks from knowledge graphs (KGs) tailored to their needs. This section outlines key parameters and their implications.
General parameters include \myNum{1} the target KG and its SPARQL endpoint, \myNum{2} the required number of dialogues, \myNum{3} the number of questions per dialogue, and \myNum{4} the LLM. {\sysName} supports both open-source and commercial LLMs, each with varying context lengths that limit input subgraph sizes. Subgraphs exceeding the LLM's context length are discarded to simplify data provision.

Regarding entity selection, {\sysName} offers two modes: automatic sampling and manual entity specification. Users can create benchmarks with specific entities by providing a text file, enabling testing with different entity types such as head, torso, and tail. Head entities typically have larger subgraphs, increasing LLM processing costs, so users must ensure they fit within the LLM’s context length. Tail entities require a sufficient number of surrounding facts to generate a dialogue. The thresholds for rare and shadowed types control which node types are included in the benchmark: a lower rare type threshold includes more node types, while a lower shadowed parent threshold excludes more parents.

Subgraph extraction parameters include \myNum{1} the number of hops from the entity. Increasing the number of hops enlarges the subgraph and LLM processing costs, resulting in questions that focus on multiple entities rather than a single center entity; \myNum{2} predicate direction, where users can select the direction of predicates (incoming, outgoing, or both); and \myNum{3} the minimum number of questions per valid dialogue, which translates to the number of unique predicates surrounding the entity.

% {\sysName} is a flexible framework that allows users to generate benchmarks from knowledge graphs (KGs) tailored to their needs. General parameters include \myNum{1} the target KG and its SPARQL endpoint, \myNum{2} the required number of dialogues, \myNum{3} the number of questions per dialogue, and \myNum{4} the LLM. 
% Regarding entity selection, {\sysName} offers two modes: automatic sampling and manual entity specification. Users can create benchmarks with specific entities by providing a text file, enabling testing with different entity types such as head, torso, and tail. Head entities typically have larger subgraphs, increasing LLM processing costs, so users must ensure they fit within the LLM’s context length. Tail entities require a sufficient number of surrounding facts to generate a dialogue. The thresholds for rare and shadowed types control which node types are included in the benchmark: a lower rare type threshold includes more node types, while a lower shadowed parent threshold excludes more parents. More details are available in \href
% {https://github.com/CoDS-GCS/Chatty-Gen/blob/main/Supplementary_material.pdf}
% % {https://gitfront.io/r/ABC/cJkrczBNnLSc/Chatbot-Resources/blob/Supplementary_material.pdf}
% {our supplementary materials}.
% % }\highlightedReply{}{R2.O1, C10}

\vspace*{-1ex}
\subsection{Evaluation Setup}

\textbf{Compared Systems:}
% {\color{blue}
We evaluate {\sysName}’s ability to generate standalone questions against Maestro \cite{maestro}, the state-of-the-art system for generating questions from KGs. We consulted the authors to reproduce Maestro’s results on DBpedia\footnote{\url{https://github.com/aorogat/Maestro}}. We extended it to work with Yago and DBLP, which involved a manual and time-consuming process to explore each KG, identify unwanted predicates, and specify suitable node labels.
% } 
\footnote{We thank the authors for reviewing our code and results for Maestro.}.
% \highlightedReply{}{R1.O3.1, C1}\shorten

\textbf{Utilized LLMs:}
We evaluated {\sysName} using various LLMs categorized into:
\myNum{I} Commercial LLMs: GPT-3.5-turbo~\cite{instructgpt}, GPT-4 \cite{gpt4}, GPT-4o \cite{gpt4o}, Gemini-1-pro \cite{geminiteam2024gemini1}, Gemini-1.5-pro \cite{geminiteam2024gemini15}. These models are accessed via API requests.
\myNum{II} Open Source LLMs: LLAMA-2-13B \cite{llama2}, LLAMA-3-8B \cite{llama3modelcard}, LLAMA-3-8B-instruct \cite{llama3modelcard}, CodeLLAMA-7B \cite{codellama}, CodeLLAMA-13B \cite{codellama}, Mistral-7B-v0.1 \cite{mistral}. 
These models are hosted on our server.

% {\color{blue}
\textbf{Four Different Real KGs:}
To evaluate {\sysName}'s scalability with large KGs from diverse domains, we tested it with four KGs: DBpedia and Yago, which cover a wide range of subjects such as people, places, and movies, and DBLP and Microsoft Academic Graph (MAG), which focus on scientific publications, authors, and citations. Their statistics are shown in \autoref{tab:stats}.
% }\highlightedReply{}{R1.O3.4, R3.O3}
\begin{table}[t]
\vspace*{-2ex}
  \caption{KG Statistics: The number of entities, and triples in Millions, The number of unique predicates and the total size in GB of each KG.
  % \highlightedReply{}{R1.O3.4, R3.O3, C8}
  }
  \label{tab:stats}
  \vspace*{-3ex}
  \begin{tabular}{ccccc}
    \toprule
    \textbf{Stat} & \textbf{DBPedia} & \textbf{YAGO} & \textbf{MAG} &\textbf{DBLP} \\
    \midrule
    \textbf{Size (GB)} &29&27&637&13\\
     \textbf{\# Predicates} &60,736& 259&178&167\\
     \textbf{\# Triples (M)} & 350& 207& 13,705& 263\\
    \textbf{\# Entities (M)} &14&12&586&18\\
   \bottomrule
\end{tabular}
\vspace*{-2ex}
\end{table}
% YAGO: Unique Predicates:  259, Num Entities:  12,598,108, Num Triples:  207029212
% DBPedia: Unique Predicates:  60736, Num Entities:  14,397,265, Num Triples:  350,677,804
% DBLP: Unique Predicates:  167 , Number of Entities:  18,306,226 , Number of Triples:  263,195,019
% MAG: Unique Predicates:  178 , Number of Entities:  568,309,304 , Number of Triples:  13705,145,395

\textbf{Computing Infrastructure:}
We use two different setups for our experiments: \myNum{I} A Linux machine with 180GB RAM and 16 CPU cores, used for deploying KGs on Virtuoso SPARQL endpoints, running {\sysName} with closed access models (via APIs), and running Maestro, and \myNum{II} A Linux machine with one Nvidia A100 40GB GPU, 32 CPU cores, and 64GB RAM, 
% {\color{blue}
used for hosting the open-source LLMs.
% }\highlightedReply{}{R1.O3.2}

\textbf{{\sysName} Implementation and Configuration:}
{\sysName} ~\footnote{Our code is available at
% ~\footnote{Upon acceptance, we will open-source our code, which is available at
\url
{https://github.com/CoDS-GCS/Chatty-Gen}
% {https://gitfront.io/r/ABC/cJkrczBNnLSc/Chatbot-Resources/}
.} is implemented using Python 3.10, the Lang chain library v0.1.4, and the OpenLLM library v0.4.44 \cite{Pham_OpenLLM_Operating_LLMs_2023} to host open-source LLMs. For SPARQL endpoints, we use Virtuoso v7.2.5.2, widely adopted for large KGs, preparing four separate Virtuoso endpoints for each KG. A standard, unmodified Virtuoso installation was used in all experiments. For node type selection, we apply a 1\% rare types threshold and a 99\% shadowed parent threshold. For entity and subgraph extraction, we use an entity batch size of 10,000, a hop count of 1, and both outgoing and incoming predicate directions.
%
% DO NOT REMOVE // Reham: This is repeated in section 6.1
% {\color{orange}{\sysName} is a configurable benchmark generation platform, allowing users to generate tailored benchmarks with different settings. Users can specify parameters such as: (i) the KG endpoint URL, (ii) the number of dialogues, (iii) the maximum number of questions per dialogue, and (iv) the LLM to use per module.}
By default, we enable our subgraph summarization and use GPT-3.5-turbo as the LLM in all experiments unless otherwise noted.\shorten

\vspace*{-1ex}
\subsection{Question Quality and Time Efficiency}
% {\color{blue}
We evaluate the self-contained questions generated by {\sysName} and Maestro based on their human-likeness, question type diversity, coverage of main KG node types, and time performance.
% }\highlightedReply{}{R1.O3.1}\shorten

\textbf{Human-Like Generated Questions:}
Maestro uses a rule-based mechanism to generate questions where the answer is the seed node. For example, the seed node for the second question in \autoref{fig:independent_example} is Harun Baraki, whose URL is \url{https://dblp.org/pid/03/8766.html}. Maestro fails to detect human-readable labels for nodes using its 
\href{https://github.com/aorogat/Maestro/blob/bad4a2f048d82233ec9d890f6427321852cf0b4d/src/main/java/knowledgeGraphs/KnowledgeGraph.java#L59}
 {\textcolor{blue}{\textit{getNodeLabel}}} function, which often uses the URL suffixes, e.g.,"baraki19", if the predicate \textit{rdfs:label} lacks an English label. In contrast, {\sysName} predicts representative entity labels and employs LLMs to generate questions that explicitly mention the seed entity. This results in more human-like, self-contained questions that cover a broader range of facts about the seed entity.
 %KEET it for CRV
 % {\color{orange}
 Consequently, questions generated by {\sysName} can be grouped to form coherent and useful dialogues, unlike Maestro's questions which often share the same answer and lack dialogue coherence.
 % }

\begin{figure}[t]
\vspace*{-2ex}
% \small
\footnotesize
	\begin{verbatim}
-------------------  Maestro: -------------------------
1. What authored by the Person "11171"? 
2. Who authored the Publication "baraki19"?
3. The authored by of the Publication "baraki19"?
4. Who is the authored by of the Publication "daun18"?
5. Is the Person "8766" the authored by of 
    the Publication "baraki19"?
------------------- Chatty-Gen: -----------------------
1. What is the primary affiliation of Jiaxin Pan?
2. Is Jiaxin Pan associated with any other affiliation?
3. Who is the author of Cooperative Situation
Awareness in Transportation?
4. In which journal was 'Trust-Region Methods
on Riemannian Manifolds.' published?
5. In which year was Cooperative Situation 
Awareness in Transportation published?
\end{verbatim}
 \vspace*{-3ex}
	\caption{Examples of questions generated from DBLP by Maestro and {\sysName}, which predicts more accurate entity labels, which helps LLMs generate human-like questions.\shorten}
	\label{fig:independent_example}
 \vspace*{-2ex}
\end{figure}

\begin{figure*}[t]
\vspace*{-2ex}
\centering
    \begin{subfigure}{.3\textwidth}
	\centering
	\includegraphics[width=\columnwidth]{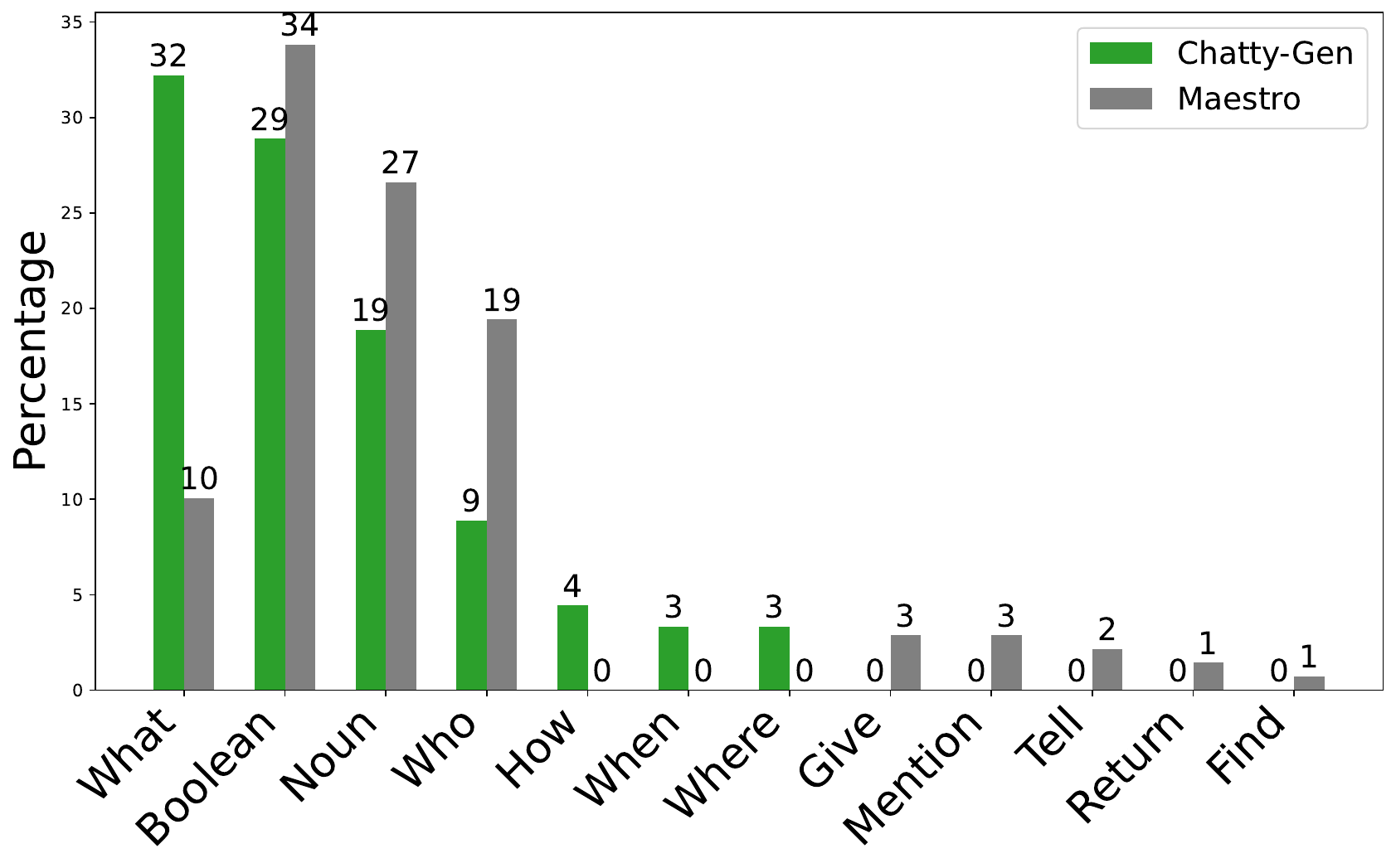}
 \vspace*{-5ex}
	\caption{DBLP KG}
	\label{fig:dblp_maestro_complexity}
	\end{subfigure}
	\hfill
         \begin{subfigure}{.3\textwidth}
   	\centering
   	\includegraphics[width=\columnwidth]{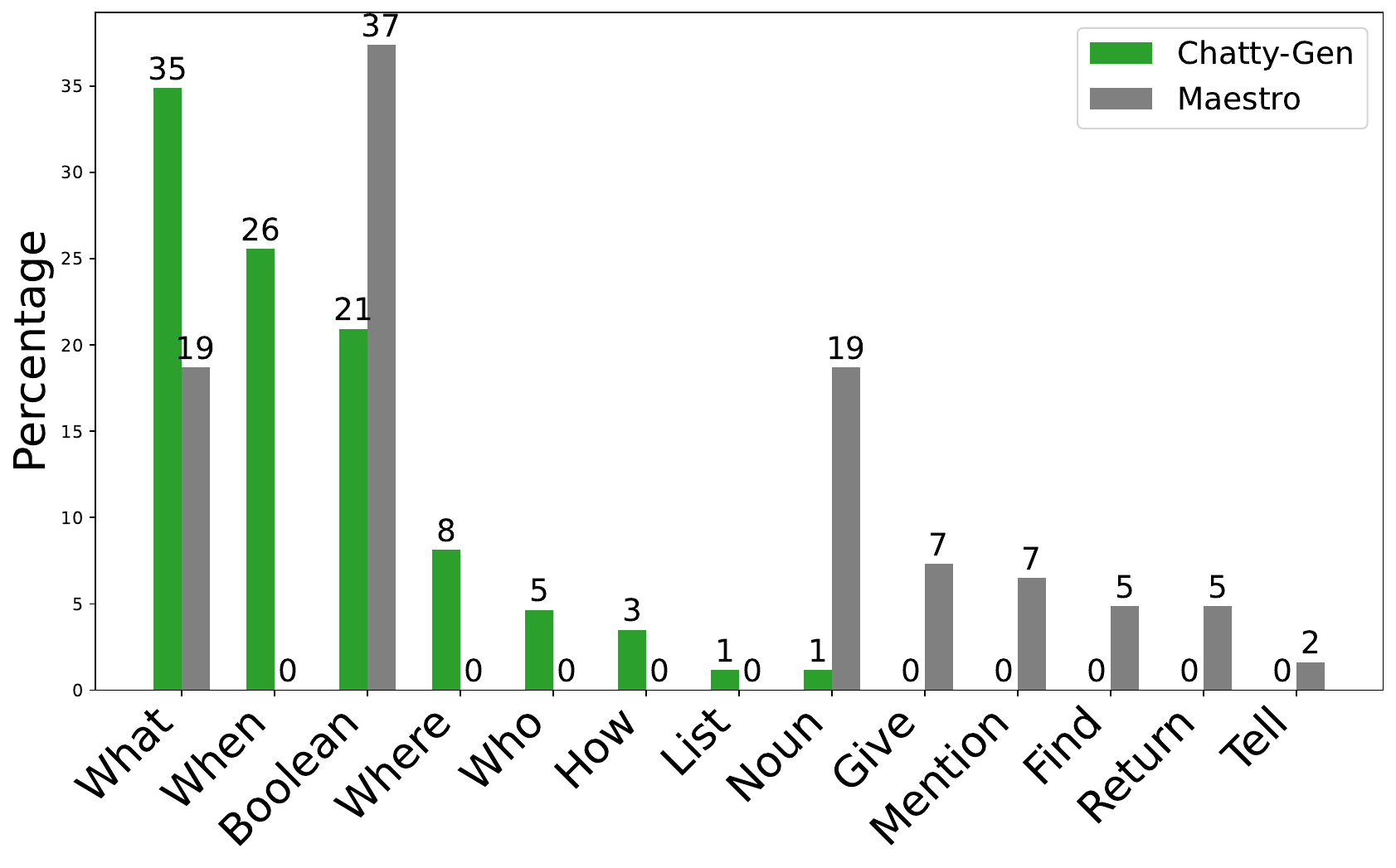}
    \vspace*{-5ex}
   	\caption{YAGO KG}
   	\label{fig:yago_maestro_complexity}
   \end{subfigure}
	\hfill
	\begin{subfigure}{.3\textwidth}
	\centering
	\includegraphics[width=\columnwidth]{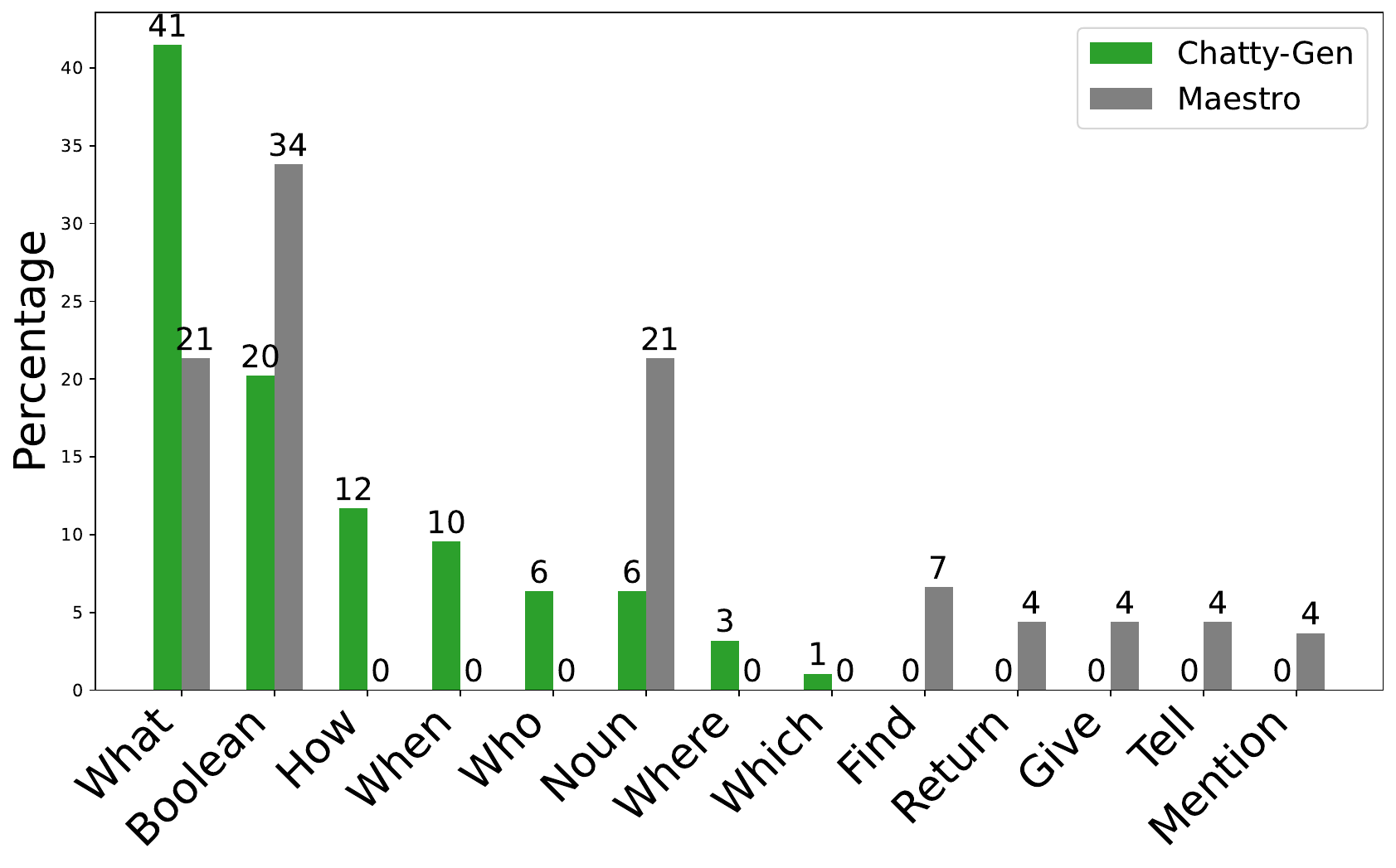}
 \vspace*{-5ex}
	\caption{DBpedia KG}
	\label{fig:maestro_complexity}
	\end{subfigure}
 \vspace*{-3ex}
	\caption{Comparison of the diversity of question types generated by Maestro and {\sysName} for the three KGs.}
	\label{fig:coverage}
 \vspace*{-2ex}
\end{figure*}

\begin{figure*}[t]
\vspace*{-1ex}
	\centering
	\begin{subfigure}{.3\textwidth}
		\includegraphics[width=\columnwidth]{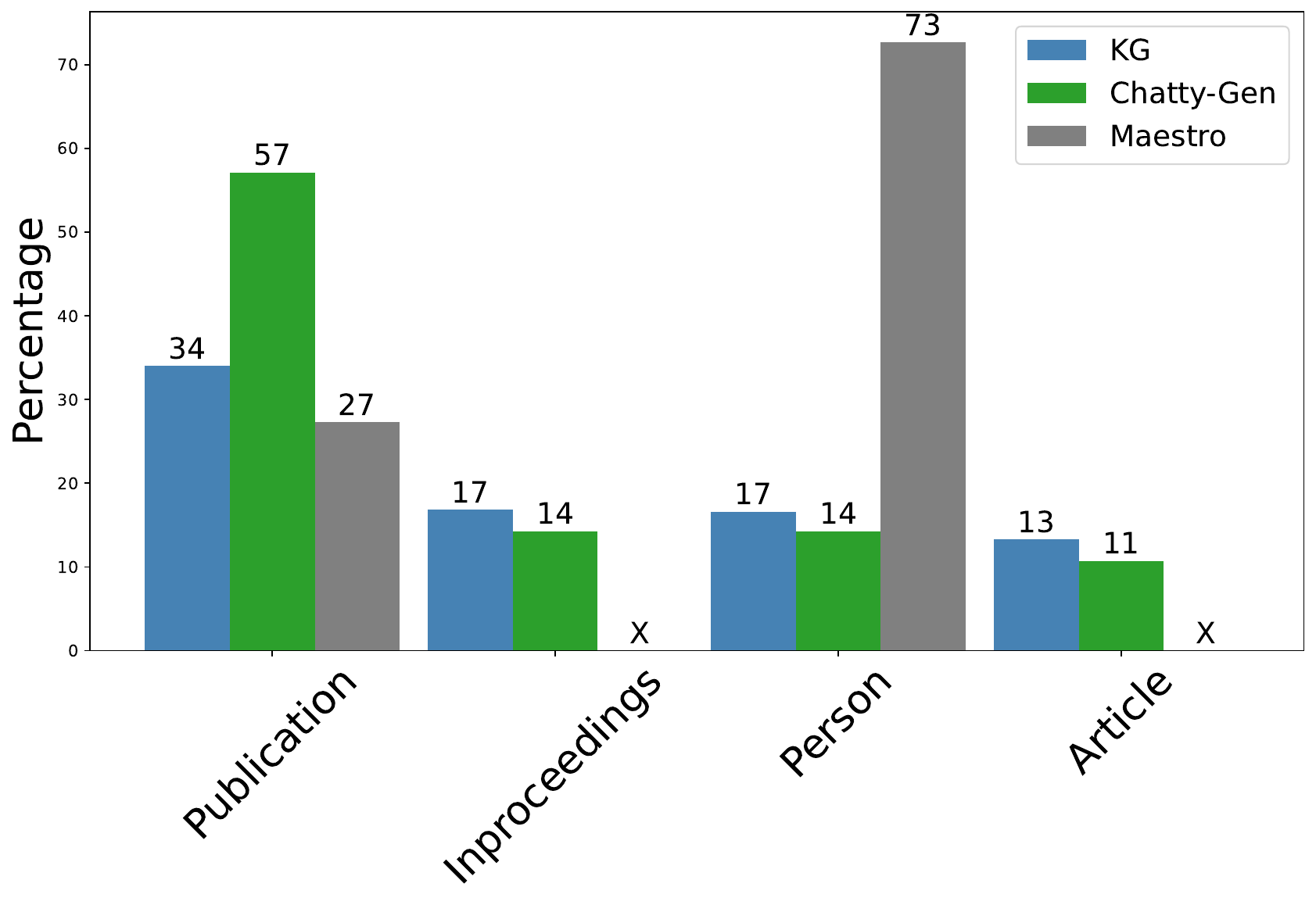}
  \vspace*{-5ex}
		\caption{DBLP KG}
		\label{fig:dblp_diversity2}
	\end{subfigure}
	\hfill
	\begin{subfigure}{.3\textwidth}
		\includegraphics[width=\columnwidth]{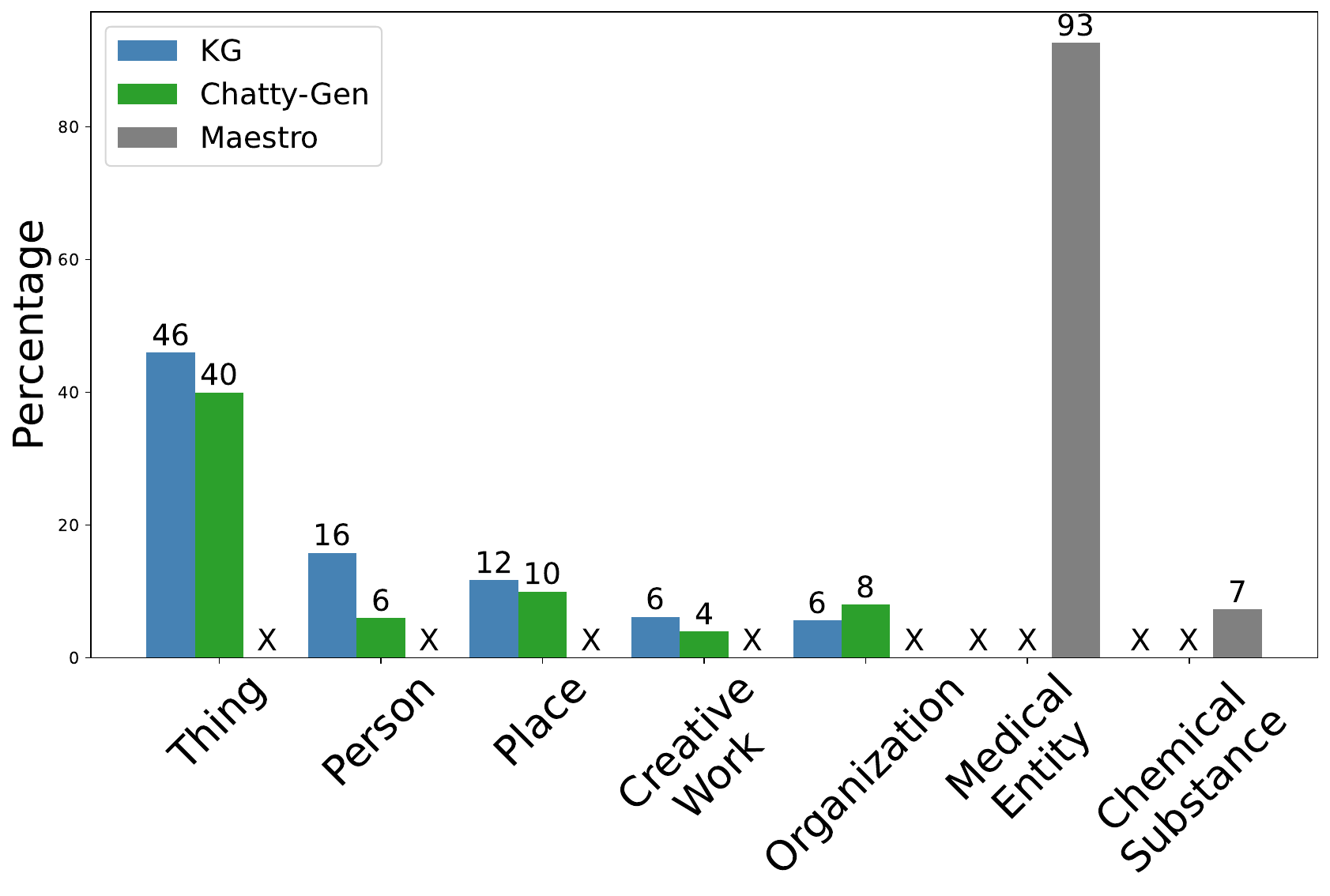}
  \vspace*{-5ex}
		\caption{YAGO KG }
		\label{fig:yago_diversity2}
	\end{subfigure}
	\hfill
	\begin{subfigure}{.3\textwidth}
		\includegraphics[width=\columnwidth]{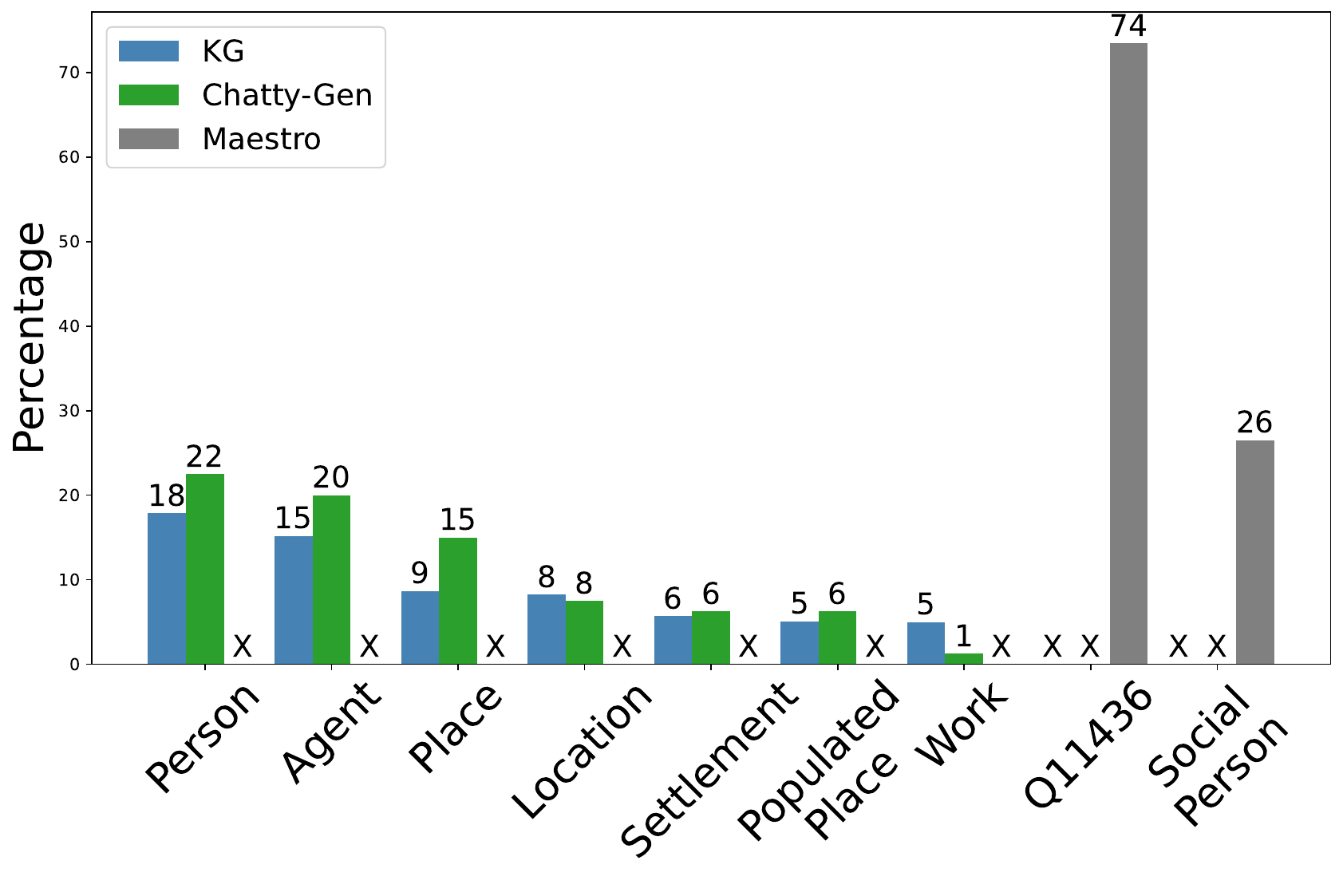}
  \vspace*{-5ex}
		\caption{DBpedia KG}
		\label{fig:dbpedia_diversity2}
	\end{subfigure}
	 \vspace*{-3ex}
	\caption{Comparison of the node-type distribution in the KG, as achieved by {\sysName} and Maestro for the selected seed entities. For the KG and {\sysName}, 'x' denotes a rare node type, whereas for Maestro, it indicates no selected seed entities.}
	\label{fig:diversity2}	
 \vspace*{-3ex}
\end{figure*}

\textbf{Diversity of Question Types:}
% {\color{blue} 
We evaluated the diversity of question types generated by {\sysName} and Maestro. \autoref{fig:coverage} shows the distribution of question types for both systems across DBLP, Yago, DBpedia, and MAG\footnote{Due to lack of space, MAG results are included in \href
{https://github.com/CoDS-GCS/Chatty-Gen/blob/main/Supplementary_material.pdf}
% {https://gitfront.io/r/ABC/cJkrczBNnLSc/Chatbot-Resources/blob/Supplementary_material.pdf}
{our supplementary materials}.}. Questions beginning with \emph{is, did, can, or has} are categorized as \emph{Boolean}, while those starting with phrases like \emph{by the agency of} or \textit{by whom} are categorized as \emph{Noun}. {\sysName} generates a wider range of question types, whereas Maestro is biased toward boolean questions, consistently producing at least $34\%$ boolean questions. In contrast, {\sysName} shows a balanced distribution with a slight preference for \emph{What} questions, typical in dialogue benchmarks~\cite{coqa, quac, csqa, convquestions}. It also maintains consistent performance across KGs, demonstrating versatility across domains.
% }\highlightedReply{}{R1.O3.3}

\textbf{Coverage of Main KG Node Types:}
We evaluated the seed node selection module in {\sysName} and Maestro to ensure that the chosen node types reflect the distribution of main node types in the KG. \autoref{fig:diversity2} illustrates the comparison of KG node type distributions between {\sysName} and Maestro\footnote{Due to lack of space, MAG results are added in \href
{https://github.com/CoDS-GCS/Chatty-Gen/blob/main/Supplementary_material.pdf}
% {https://gitfront.io/r/ABC/cJkrczBNnLSc/Chatbot-Resources/blob/Supplementary_material.pdf}
{our supplementary materials}.}. Node types were obtained by querying the KG endpoint for each chosen entity and calculating the percentage of each type. To enhance clarity, KG types with less than $5\%$ representation were excluded from the figure.
%
% {\color{blue} 
{\sysName} accurately represents the main node types in the KG, while Maestro is not optimized to select enities based on most significant types.
% }\highlightedReply{}{R1.O3.3}
In {\sysName}, most node types closely match the KG distribution. However, due to some nodes having multiple types, a single seed may contribute to multiple categories. In contrast, Maestro did not include many seeds from the most representative types in the KG. For example, the Person type, constituting $18\%$ of entities in the DBpedia KG, is absent from Maestro's benchmarks. Instead, Maestro generates a high percentage of rare types, such as Social Person (representing $26\%$ of its benchmarks).\shorten

\textbf{Time Performance in Question Generation:}
This experiment analyzes the time taken by Maestro and \texttt{\sysName} to generate benchmarks of $100$ questions for different KGs. \autoref{tab:maestro_time} presents the time taken by both systems
% {\color{blue},
It demonstrates the efficiency of {\sysName} compared to Maestro, which takes hours to generate benchmarks from large KGs.
% }\highlightedReply{}{R1.O3.3}
%
Maestro first requires traversing the entire KG to extract all predicates along with their contexts, defined as (predicate, subject\_type, object\_type), which are stored in a PostgreSQL Database for subsequent access. The benchmark generation process begins after retrieving the required number of seed entities from the database. For DBLP and YAGO, all predicates were included, while for DBpedia, the process was halted at $18,211$ out of $994,592$ predicates due to memory limitations, requiring several days to complete. Maestro's overall processing time is dominated by this preprocessing step, which scales proportionally with the number of predicates.
In contrast, {\sysName} dynamically utilizes information from the entire KG in real-time and completes the end-to-end generation process in minutes, regardless of the KG size. This significant reduction in processing time can be attributed to our effective seed sampling process, as discussed in \autoref{sec:seed}.
% {\color{blue}
In this experiment, {\sysName} uses a commercial LLM, which incurs additional overhead from remote API requests—unlike Maestro. Despite this overhead, {\sysName} still shows faster performance.
% }\highlightedReply{}{R1.O3.3, C12}\shorten

\begin{table}[t]
% \vspace*{-1ex}
  \caption{End to end time performance comparison between {\sysName} and Maestro in hours. Maestro is affected by number of predicates and size of the KG.}
  \label{tab:maestro_time}
  \vspace*{-3ex}
  \begin{tabular}{cccc}
    \toprule
    \textbf{KG} & \textbf{Maestro} & \textbf{{\sysName}} \\
    \midrule
    \textbf{DBpedia} & 30.77 & 0.17 \\
    \textbf{YAGO} & 5.20  & 0.10 \\
    \textbf{MAG} & 5.38 & 0.12\\
    \textbf{DBLP} &  0.12  & 0.12\\
  \bottomrule
\end{tabular}
\vspace*{-1ex}
\end{table}
\begin{table*}[t]
\small
	\vspace*{-2ex}
	\caption{Evaluating the performance of our multi-stage approach vs the single prompt approach using commercial and open-source LLMs and two KGs, DBLP and YAGO. \texttt{E} denotes error, and \texttt{S} denotes successful results, Multi-LLM-1: (Question-triple and Dialogue Generation: LLAMA-3-8b-inst + Answer Queries Generation: CodeLLAMA-13b), Multi-LLM-2: (Question-triple Generation: GPT-4, Dialogue Generation: LLAMA-3-8b-inst, + Answer Queries Generation: CodeLLAMA-13b).}
	\label{tab:openllm}
 \vspace*{-3ex}
	\begin{tabular}{c|c|cccccccc|cccccccc}
		% \toprule
  \hline
		 & &  \multicolumn{8}{c|}{\textbf{YAGO}} & \multicolumn{8}{c}{\textbf{DBLP}} \\
   % \midrule
   \hline
		 \begin{sideways}\textbf{Approach}\end{sideways}& \textbf{LLM} & 
        
        % \begin{sideways}\textbf{Num\_seeds}\end{sideways} & 
        \begin{sideways}\textbf{Success rate \%}\end{sideways} & \begin{sideways}\textbf{Dialogue-S}\end{sideways} & \begin{sideways}\textbf{Question-E}\end{sideways} & \begin{sideways}\textbf{SPARQL-E}\end{sideways}  & \begin{sideways}\textbf{Dialogue-E}\end{sideways} &  \begin{sideways}\textbf{Parsing-E}\end{sideways} & \begin{sideways}\textbf{Time(mins)}\end{sideways} & \begin{sideways}\textbf{\# Tokens M (\$)}\end{sideways} &
        
        % \begin{sideways} \textbf{Num\_seeds}\end{sideways} & 
        \begin{sideways}\textbf{Success rate \%}\end{sideways} & 
        \begin{sideways} \textbf{Dialogue-S}\end{sideways} & \begin{sideways}\textbf{Question-E}\end{sideways} & \begin{sideways}\textbf{SPARQL-E}\end{sideways}  & \begin{sideways}\textbf{Dialogue-E}\end{sideways} & \begin{sideways}\textbf{Parsing-E}\end{sideways} & \begin{sideways}\textbf{Time(mins)}\end{sideways} & \begin{sideways}\textbf{\# Tokens M (\$)}\end{sideways}\\
		% \midrule
            \hline
	    \multirow{13}{*}{
            \rotatebox[origin=c]{90}{\textbf{Our Multi-stages (3 prompts)}}}
            & \textbf{GPT-4o} & 100 & 20 & 0 & 0 & 0 & 0 & 8 & 0.04(0.32) & 95 & 20 & 0 & 0 & 1& 0 & 6 & 0.05(0.38) \\
            & \textbf{GPT-4} & 95 & 20 & 0 & 0 & 1 & 0 & 13 & 0.05(0.71) & 100 & 20 & 0 & 0 & 0 & 0 & 11 & 0.05(0.78) \\
	    & \textbf{GPT-3.5} & \cellcolor{lightgray} 91 &  \cellcolor{lightgray}20 &  1 & 0 & 1 & 0 &  \cellcolor{lightgray}6 &  \cellcolor{lightgray}0.05(0.04) &  \cellcolor{lightgray}95 &  \cellcolor{lightgray}20 & 1 & 0 & 0 & 0 &  \cellcolor{lightgray}7 &  \cellcolor{lightgray}0.06(0.05) \\
            & \textbf{Gemini-1-pro} &  71 &  20 & 0 & 2 & 0 & 6 & 6 & 0.06(0.01) & 67 &  20 & 5 & 0 & 0 & 5 & 5.2 & 0.09(0.02)\\
            & \textbf{Gemini-1.5-pro} & 22 & 20 & 0 & 1 & 0 & 71 & 20.5 & 0.18(0.35) & 41 & 20 & 0 & 0 & 1 & 28 & 14.5 & 0.12(0.24)\\
            \cline{2-18}
            & \textbf{{LLAMA-3-8b}} & 13 & 20 & 2 & 124 & 9 & 0 & 84 & 0.36 & 20 & 20 & 2 & 47 & 23 & 8 & 71 & 0.33 \\
            & \textbf{LLAMA-3-8b-inst} &  \cellcolor{lightgray}41 &  \cellcolor{lightgray}20 & 0 & 16 & 0 & 13 & 37 & 0.13 &  \cellcolor{lightgray}14 &  \cellcolor{lightgray}20 & 0 & 106 & 0 & 16 & 89 & 0.39 \\
            & \textbf{LLAMA-2-13b} & 5 & 20 & 5 & 325 & 10 & 1 & 270 & 0.79 & 1 & 5 & 32 & 426 & 7 & 22 & 392 & 1.30 \\
            & \textbf{CodeLLAMA-7b} & 7 & 20 & 3 & 162 & 74 & 16 & 234 & 0.92 & 1 & 5 & 144 & 156 & 44 & 145 & 356 & 0.16 \\
	    & \textbf{CodeLLAMA-13b} &  \cellcolor{lightgray}83 &  \cellcolor{lightgray}20 & 0 & 0 & 4 & 0 &  \cellcolor{lightgray}31 & 0.08 &  \cellcolor{lightgray}63 &  \cellcolor{lightgray}20 & 3 & 5 & 2 & 1 &  \cellcolor{lightgray}37 & 0.01 \\
            & \textbf{Mistral-7b-v0.1} &  \cellcolor{lightgray}17 &  \cellcolor{lightgray}20 & 2 & 88 & 3 & 0 & 70 & 0.25 & 3 & 15 & 5 & 323 & 5 & 14 & 189 & 0.90 \\
             \cline{2-18}
            & \textbf{Multi-LLM-1} &  \cellcolor{lightgray}100 &  \cellcolor{lightgray}20 & 0 & 0 & 0 & 0 & 18 &  \cellcolor{lightgray}0.06 &  \cellcolor{lightgray}83 &  \cellcolor{lightgray}20 & 0 & 4 & 0 & 0 & 20 &  \cellcolor{lightgray}0.07\\
            & \textbf{Multi-LLM-2} & \cellcolor{lightgray}91 & \cellcolor{lightgray}20 & 0 & 0 & 0 & 2 & 18 &  \cellcolor{lightgray}0.06(0.28) & \cellcolor{lightgray}71 & \cellcolor{lightgray}20 & 2 & 4 & 0 & 2 & 23 & \cellcolor{lightgray}0.10(0.82)\\
	    % \midrule
            \hline
	    \multirow{11}{*}{
            \rotatebox[origin=c]{90}{\textbf{single prompt}}}
            % \textbf{One-stage}}
            
             & \textbf{GPT-4o} & 38 & 20 & 11 & 0 & 5 & 15 & 9 & 0.03(0.32) & 8 & 20 & 52 & 0 & 8 & 144 & 59 & 0.15(1.30)\\
             & \textbf{GPT-4} & 40 & 20 & 5 & 1 & 1 & 23 & 17 & 0.04(0.73) & 21 & 20 & 5 & 1 & 2 & 66 & 46 & 0.09(1.86)\\
             & \textbf{GPT-3.5} &  \cellcolor{lightgray}5 &  \cellcolor{lightgray}19 & 2 & 0 & 8 & 242 &  \cellcolor{lightgray}53 &  \cellcolor{lightgray}0.36(0.32) &  \cellcolor{lightgray}8 &  \cellcolor{lightgray}20 & 25 & 0 & 5 & 119 &  \cellcolor{lightgray}50 &  \cellcolor{lightgray}0.25(0.24)\\
            & \textbf{Gemini-1-pro} & 56 & 20 & 0 & 10 & 4 & 1 & 3 & 0.03(0.01) &  10 & 20 & 130 & 0 & 12 & 6 & 21 & 0.30(0.07)\\
            & \textbf{Gemini-1.5-pro} & 14 & 20 & 59 & 4 & 0 & 5 & 16 & 0.12(0.27) & 6 & 20 & 28 & 0 & 3 & 10 & 51.5 & 0.43(0.9)\\
            \cline{2-18}
            & \textbf{LLAMA-3-8b} & 0 & 0 & 44 & 107 & 158 & 92 & 251 & 1 & 0 & 2 & 170 & 23 & 55 & 452 & 286 & 1.23\\
            & \textbf{LLAMA-3-8b-inst} &  \cellcolor{lightgray}9 &  \cellcolor{lightgray}20 & 10 & 13 & 59 & 43 & 72 & 0.26 &  \cellcolor{lightgray}7 &  \cellcolor{lightgray}20 & 51 & 6 & 44 & 71 & 90 & 0.38\\
            & \textbf{LLAMA-2-13b} & 0 & 0 & 40 & 216 & 87 & 401 & 403 & 1 & 0 & 0 & 53 & 39 & 18 & 827 & 479 & 1.22\\
            & \textbf{CodeLLAMA-7b} & 0 & 0 & 72 & 87 & 196 & 344 & 314 & 1.07 & 0 & 1 & 203 & 45 & 30 & 604 & 313 & 1.25\\
            & \textbf{CodeLLAMA-13b} &  \cellcolor{lightgray}1 & \cellcolor{lightgray} 11 & 78 & 29 & 579 & 168 &  \cellcolor{lightgray}443 & 1.07 &  \cellcolor{lightgray}1 &  \cellcolor{lightgray}5 & 188 & 11 & 91 & 677 &  \cellcolor{lightgray}459 & 1.26\\
            & \textbf{Mistral-7b-v0.1} &  \cellcolor{lightgray}0 & \cellcolor{lightgray} 0 & 53 & 93 & 420 & 223 & 261 & 0.98 & 0 & 1 & 208 & 34 & 118 & 528 & 257 & 1.20\\
            \hline
	\end{tabular}
 \vspace*{-3ex}
\end{table*}

\vspace*{-2.5ex}
\subsection{Consistent Performance with Diverse LLMs}
This experiment compares our multi-stage approach to a single-prompt approach using a diverse set of commercial and open-source LLMs with two KGs: YAGO and DBLP. Open-source LLMs offer cost-effective solutions for long-term projects and eliminate the need to send private data to external APIs. Consequently, we optimized our platform to achieve performance with open-source LLMs comparable to our performance with commercial LLMs.

\textbf{Open-Source vs. Commercial LLMs:}
We tested representative samples of commercial and open-source LLMs. \autoref{tab:openllm} summarizes their performance across different KGs.
We measured the following metrics: dialogues generated successfully (\texttt{Dialogues-S}), \texttt{Success Rate} (\texttt{Dialogues-S}/No. of visited Seed Nodes), validation errors (\texttt{Questions-E}, \texttt{Triples-E}, \texttt{Dialogue-E}), parsing errors (\texttt{Parsing-E}), time in minutes, and the total number of tokens in millions. We also reported the equivalent dollar cost based on the current pricing schema. Each LLM was tasked with generating benchmarks of 20 dialogues, each with up to 5 questions. We set a limit of 500 seed entities for the multi-stage approach and 1000 for the one-stage approach.
% {\color{blue}
%  More errors mean increased hallucination and lower success rates.
The hallucination rate is indicated by success and error rates across multiple stages, i.e., more errors and lower success rates mean a higher ratio of hallucinations. To compare open-source and commercial LLMs on metrics unaffected by infrastructure, we focus on success rate, error rates, and token count.
% }\highlightedReply{}{R1.O3.2, C6, C9}
%
For open-source LLMs, our multi-stage approach significantly improves the success rate compared to a single prompt approach. Only one model (\texttt{LLAMA-3-8b-inst}) successfully generated all dialogues using the single prompt approach, while all models succeeded with the multi-stage approach.
For commercial LLMs, despite requiring only a single prompt compared to three in the multi-stage approach, the multi-stage method achieves significantly higher success rates. For example, \texttt{GPT-3.5-turbo} achieved success rates of 91\% and 95\% for YAGO and DBLP, respectively, using the multi-stage approach, compared to 5\% and 8\% using the single-prompt approach. Our approach also results in lower cost and time with \texttt{GPT-3.5-turbo} due to processing fewer nodes and avoiding the need to restart tasks caused by hallucinations.\shorten

\textbf{Multi-Stage with Multiple LLMs (Multi-LLMs):}
We tested two variations of {\sysName} with Multi-LLMs, combining the best-performing open-source and commercial models from each stage:
\myNum{1} \texttt{Multi-LLM-1}: Uses \texttt{LLAMA-3-instruct (8B)} for question-triple and dialogue generation, and \texttt{CodeLLAMA (13B)} for SPARQL queries (open-source only).
\myNum{2} \texttt{Multi-LLM-2}: Uses \texttt{GPT-4} for question-triple generation, \texttt{LLAMA-3-instruct (8B)} for dialogue generation, and \texttt{CodeLLAMA (13B)} for SPARQL queries (uses \texttt{GPT-4} for one stage).
The computing resources of commercial models, such as GPT-4o and Gemini, are intuitively more powerful than our server used for running the open-source LLMs. Despite this fact, our Multi-LLM variations achieve performance comparable to the commercial models. \texttt{Multi-LLM-1} uses only open-source models, while \texttt{Multi-LLM-2} uses \texttt{GPT-4} for only one stage, demonstrating the cost-effectiveness of our approach.
In conclusion, our multi-stage approach in {\sysName} effectively improves the performance of open-source and commercial LLMs in dialogue benchmark generation tasks. \texttt{\sysName} offers a more efficient and cost-effective solution by dividing complex tasks into sub-tasks and leveraging the strengths of different models. The flexibility to use different models in different stages allows for customization and improved success rates across a broader range of LLMs.\shorten

\begin{figure}[t]
% \vspace*{-2ex}
  \centering
  \includegraphics[width=\columnwidth]{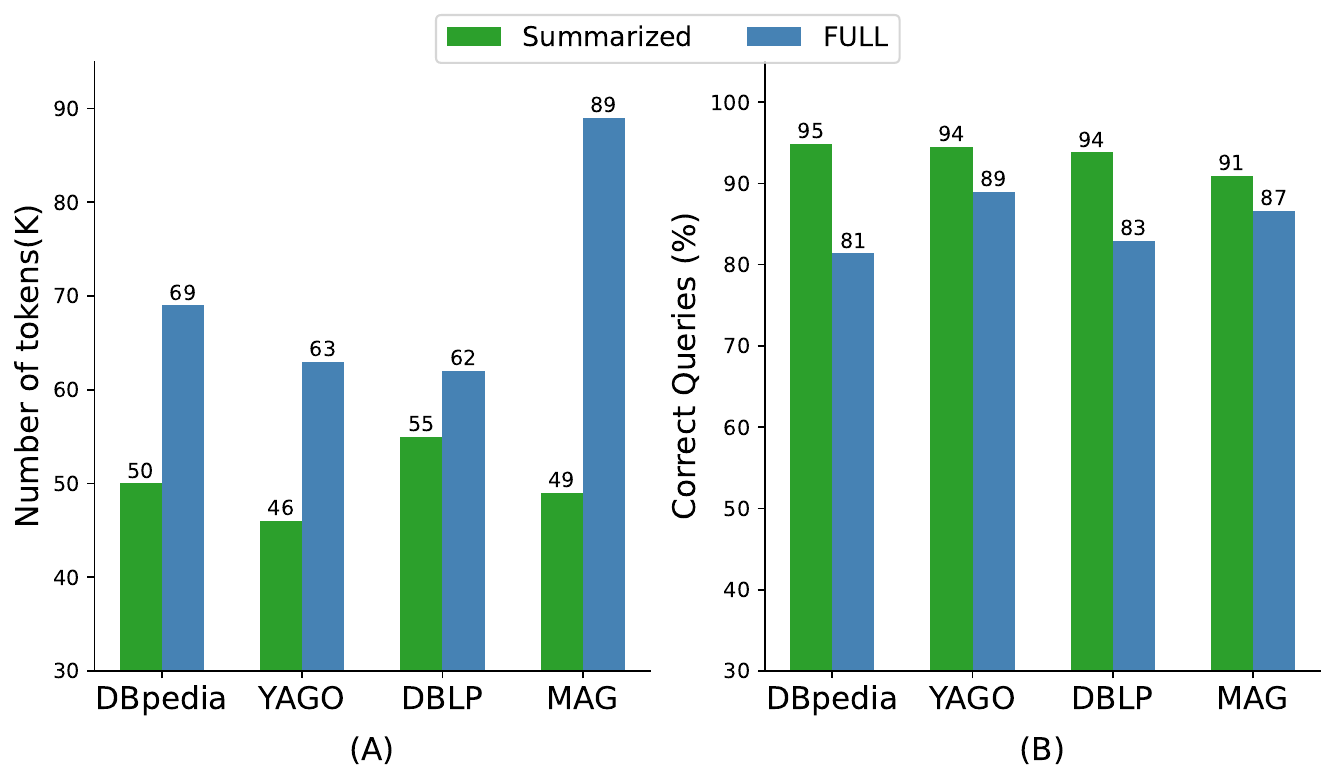}
  \vspace*{-6ex}
\caption{Comparing the performance of {\sysName} when using Full and Summarized subgraphs. (A) shows the total number of tokens used to generate the benchmark, and (B) shows the percentage of correctly generated SPARQL queries. 
}
\label{fig:ser}
% \vspace*{-2ex}
\end{figure}

\textbf{Serialization for LLM Prompts:}
This experiment evaluates {\sysName} in two settings: using full subgraphs and summarized subgraphs. The analysis focuses on the number of tokens consumed (
% {\color{blue} 
the lower, the better
% }
) and the generation of correct SPARQL queries (
% {\color{blue} 
the higher, the better
% }
), as depicted in Figure~\ref{fig:ser}.
We used GPT-3.5-turbo as the LLM. Compared to the full subgraph approach, our summarized subgraph approach reduces the total number of tokens while achieving a higher ratio of correct SPARQL queries per KG. Both approaches show comparable time performance.\shorten
% \vspace*{-2ex}
% {\color{blue}
%\section{Limitations, Future Work, and Use Case}
\section{Discussion: Use Case and Limitations}
\noindent \textbf{Use case:}
Human-generated dialogue benchmarks are both time-consuming and expensive. For instance, creating the widely used ConvQuestions benchmark~\footnote{\url{https://convex.mpi-inf.mpg.de}}~\cite{convquestions} through Amazon Mechanical Turk (AMT) cost thousands of euros. The task involved generating five natural questions and their corresponding answers (via web search) for five entities of the participant choice across topics (entity types), such as books, movies, music, soccer, and TV series. Seventy participants were given 3 hours each per task. 
The payment was 25 euros per task.
This use case highlights the cost-effectiveness of {\sysName} compared to ConvQuestions.\shorten

Following common practice in using LLMs to evaluate the quality of NLP tasks ~\cite{llm_judge}, we used LLMs as judges in our use-case. Specifically, we tested {\sysName} with GPT-4o and a multi-LLM setup (LLAMA-3-8b-inst and CodeLLAMA-13). These are the top-performing setup as detailed in Table~\ref{tab:openllm}. Hence, we utilized Gemini 1.5~\footnote{The prompt used is available in \href
{https://github.com/CoDS-GCS/Chatty-Gen/blob/main/Supplementary_material.pdf}
% {https://gitfront.io/r/ABC/cJkrczBNnLSc/Chatbot-Resources/blob/Supplementary_material.pdf}
{our supplementary materials}.} to evaluate fluency, clarity, and variety by comparing sets of five questions from both benchmarks. To avoid positional bias as indicated in~\cite{llm_judge}, we presented each dialogue pair to the \textit{Judge} twice in different orders. If the evaluations differed, the outcome was marked as a \textbf{tie}.
\autoref{tab:usecase} shows tie rates of 70\% and 80\% with {\sysName} using GPT-4o and Multi-LLM-1, respectively. ConvQuestions was favored in up to 25\% of cases, while {\sysName} with GPT-4o won in 5\%. In the 25\% of cases where {\sysName} was not selected, the quality remained comparable; however, the \textit{Judge} was required to make a choice~\footnote{Due to space constraints, samples are provided in \href
{https://github.com/CoDS-GCS/Chatty-Gen/blob/main/Supplementary_material.pdf}
% {https://gitfront.io/r/ABC/cJkrczBNnLSc/Chatbot-Resources/blob/Supplementary_material.pdf}
{our supplementary materials}.}.
%
% In conclusion, generating dialogues for 20 entities in ConvQuestions costs 100 euros and takes at least three hours, while {\sysName} can generate dialogues of comparable quality in about 15 minutes for just \$0.27 USD. 
In conclusion, generating dialogues for 20 entities in ConvQuestions costs approximately 100 euros and requires at least three hours with four participants. In contrast, {\sysName} can produce dialogues of comparable quality in about 15 minutes at a cost of just \$0.27 USD.
% \highlightedReply{}{R1.O4, R3.O2}

\begin{table}[t]
% \vspace*{-2ex}
    \centering
    % \caption{Comparison of question quality between human-generated questions from ConvQuestions and automatically generated questions by {\sysName} using different LLMs. }
    \caption{Comparison between ConvQuestions and {\sysName} dialogues using different LLMs, with \textit{tie} indicating indistinguishable quality, and the favored ratio shown otherwise.\shorten}
    \vspace*{-1ex}
    \label{tab:usecase}
    \begin{tabular}{cccc}
    \toprule
        \textbf{Model} &  \textbf{Tie} & \textbf{ConvQuestions} & \textbf{Chatty-Gen} \\
        \midrule
         \textbf{GPT-4o}& 70\% & 25\% & 5\%\\
         \textbf{Multi-LLM-1} & 80\% & 20\% & 0\%\\
         % \textbf{CodeLLAMA-13b} & 25\% & 75\% & 0 \\
         % ConvQuestions & 25\\
         % Chatty-Gen& 5\\
         \bottomrule
    \end{tabular}
    % \vspace*{-2ex}
\end{table}
% Values
% Gpt-4o:   "Tie": 14, "ConvQuestions": 5, "Chatty_Gen": 1
% Mulit-LLM1: Tie:  12, Chatty-Gen:  0, Conv-Questions:  7
% Multi-LLM1 - 7q: Tie:  16, Chatty-Gen:  1, Conv-Questions:  3
% CodeLLAMA13b: Tie:  4, Chatty-Gen:  0, Conv-Questions:  16

\noindent \textbf{Limitations and Future work:}
{\sysName} prioritizes cost-effective dialogue generation, which may lead to discarding entities when hallucinations occur. The presence of similar entities in large KGs encourages {\sysName} to discard entities instead of fixing errors, which exceed a certain cost. This approach can limit users seeking specific entities, even at additional cost and time. A potential future extension is a correction module that addresses errors instead of discarding entities. This module would handle various errors, including incorrect triples and misplaced questions.
Moreover, most existing question answering systems (QAS) for KGs, such as KGQAn\cite{kgqan}, NSQA\cite{nsqa} and EDGQA\cite{EDGQA}, lack dialogue support. {\sysName} poses research opportunities to develop interactive QAS for KGs with dialogue capabilities and human-like answers. 
% }\highlightedReply{}{R3.O1}

\section{Related Work}
Creating dialogue benchmarks from KGs is an emerging area within AI benchmark development. Previous efforts have focused on document-based conversational datasets like CoQA \cite{coqa} and QuAC \cite{quac}. Early efforts in dialogue benchmark creation on KGs include manually curated datasets, such as QALD \cite{qald9}, LCQuAD \cite{lcquad, lcquad2} on DBpedia \cite{dbpedia}, DBLP-QuAD \cite{dblpquad} on DBLP KG, and WEBQUESTIONS \cite{webquestions} on Freebase KG. More recently, benchmarks like Head-to-Tail \cite{head_to_tail} and Maestro \cite{maestro} focus mainly on generating standalone and self-contained questions from KGs. These efforts did not address the need for dialogue benchmark generation from KGs, which has challenging requirements. CSQA is a dialogue benchmark generated from the Wikidata KG and involves semi-automated processes, including manual KG exploration and template creation, which still demand massive human-intensive work. Our system, {\sysName}, is the first to address the creation of dialogue benchmarks specifically from KGs, leveraging the capabilities of LLMs to generate contextually rich questions directly from KG subgraphs. 

The prompt-based approaches, such as Chain of Thought\cite{cot} and least-to-most prompting \cite{least_to_most}, are used to achieve state-of-the-art performance in tasks like Text-to-SQL \cite{dinsql, text_sql_cot} and code generation \cite{codegeneration, codegeneration2}. Generally, commercial LLMs, such as GPT-4~\cite{gpt4} and Gemini~\cite{gemini}, tend to achieve better results across different tasks. However, they can be costly for tasks involving frequent or lengthy prompts. This has led to a growing focus on improving the performance of open-source models,  like LLama and Mistral\cite{maestro} to achieve comparable performance to commercial-LLMs in some tasks, such as dialogue state tracking \cite{DST}. Unlike these systems, we propsoed a multi-stage approach that helps to improve the over all performance of both commercial and open-source LLMs plus integrating different open-source LLMs into {\sysName} to achieve comparable performance to the commercial LLMs, such as GPT-4.\shorten

% \vspace*{-3ex}
\section{Conclusion}
This paper introduced {\sysName}, a novel RAG-based platform for automating dialogue benchmark generation from KGs. Our multi-stage approach with zero-shot learning helps {\sysName} efficiently work across diverse KGs. {\sysName} leverages LLMs to predict entity labels and generate contextually rich questions. Hence, our platform achieves significant advancements in dialogue benchmark quality. Our evaluation across DBpedia, Yago, and DBLP KGs demonstrate {\sysName}'s seamless integration with commercial and open-source LLMs. {\sysName} consistently delivers comparable performance across these LLMs while reducing processing time and cost. {\sysName} outperforms the state-of-the-art system in question quality and processing efficiency.

Hence, {\sysName} presents a versatile and cost-efficient solution for generating high-quality dialogue benchmarks tailored to specific domains from KGs. This work establishes the foundation for further advancements in automated benchmark systems, enhancing their applicability in domains requiring structured knowledge interaction and evaluation.\shorten

\balance
\bibliographystyle{ACM-Reference-Format}
\bibliography{References}

\end{document}